%% file: main.tex
\documentclass[10pt,twocolumn,letterpaper]{article}

\usepackage{cvpr}              

\makeatletter
\def\blfootnote{\xdef\@thefnmark{}\@footnotetext}
\makeatother

\input{preamble}
\definecolor{cvprblue}{rgb}{0.21,0.49,0.74}
\usepackage[pagebackref,breaklinks,colorlinks,citecolor=cvprblue]{hyperref}
\usepackage{overpic} 
\usepackage{multirow}
\usepackage{utfsym}
\usepackage{makecell}

\def\paperID{32213} 
\def\confName{CVPR}
\def\confYear{2026}

\title{SelfHVD: Self-Supervised Handheld Video Deblurring}

\author{
Honglei Xu, Zhilu Zhang$^*$, Junjie Fan, Xiaohe Wu, Wangmeng Zuo \\
Harbin Institute of Technology\\
{\tt\small cshongleixu@gmail.com, cszlzhang@outlook.com,} {\tt\small \{zhou93108, csxhwu, cswmzuo\}@gmail.com}
}

\begin{document}
\maketitle

\blfootnote{$^*$ Corresponding author.}

\input{sec/0_abstract}
\input{sec/1_intro}
\input{sec/2_relatedwork}
\input{sec/3_method}

\input{sec/4_experiment}
\input{sec/5_ablation}
\input{sec/6_conclusion}
\clearpage
\input{sec/7_acknowledgements}
{
    \small
    \bibliographystyle{ieeenat_fullname}
    \bibliography{main}
}
\input{sec/X_suppl}


\end{document}

%% file: sec/0_abstract.tex
\begin{abstract}
Shooting video with handheld shooting devices often results in blurry frames due to shaking hands and other instability factors. Although previous video deblurring methods have achieved impressive progress, they still struggle to perform satisfactorily on real-world handheld video due to the blur domain gap between training and testing data. To address the issue, we propose a self-supervised method for handheld video deblurring, which is driven by sharp clues in the video. First, to train the deblurring model, we extract the sharp clues from the video and take them as misalignment labels of neighboring blurry frames. Second, to improve the deblurring ability of the model, we propose a novel Self-Enhanced Video Deblurring (SEVD) method to create higher-quality paired video data. Third, we propose a Self-Constrained Spatial Consistency Maintenance (SCSCM) method to regularize the model, preventing position shifts between the output and input frames. Moreover, we construct synthetic and real-world handheld video datasets for handheld video deblurring. Extensive experiments on these and other common real-world datasets demonstrate that our method significantly outperforms existing self-supervised ones. The code and datasets are publicly available at \url{https://cshonglei.github.io/SelfHVD}.
\end{abstract}

%% file: sec/1_intro.tex
\section{Introduction}
\label{sec:intro}

\begin{figure}[t]
    \centering
    \begin{subfigure}[b]{1\linewidth}
        \centering
        \includegraphics[width=\linewidth]{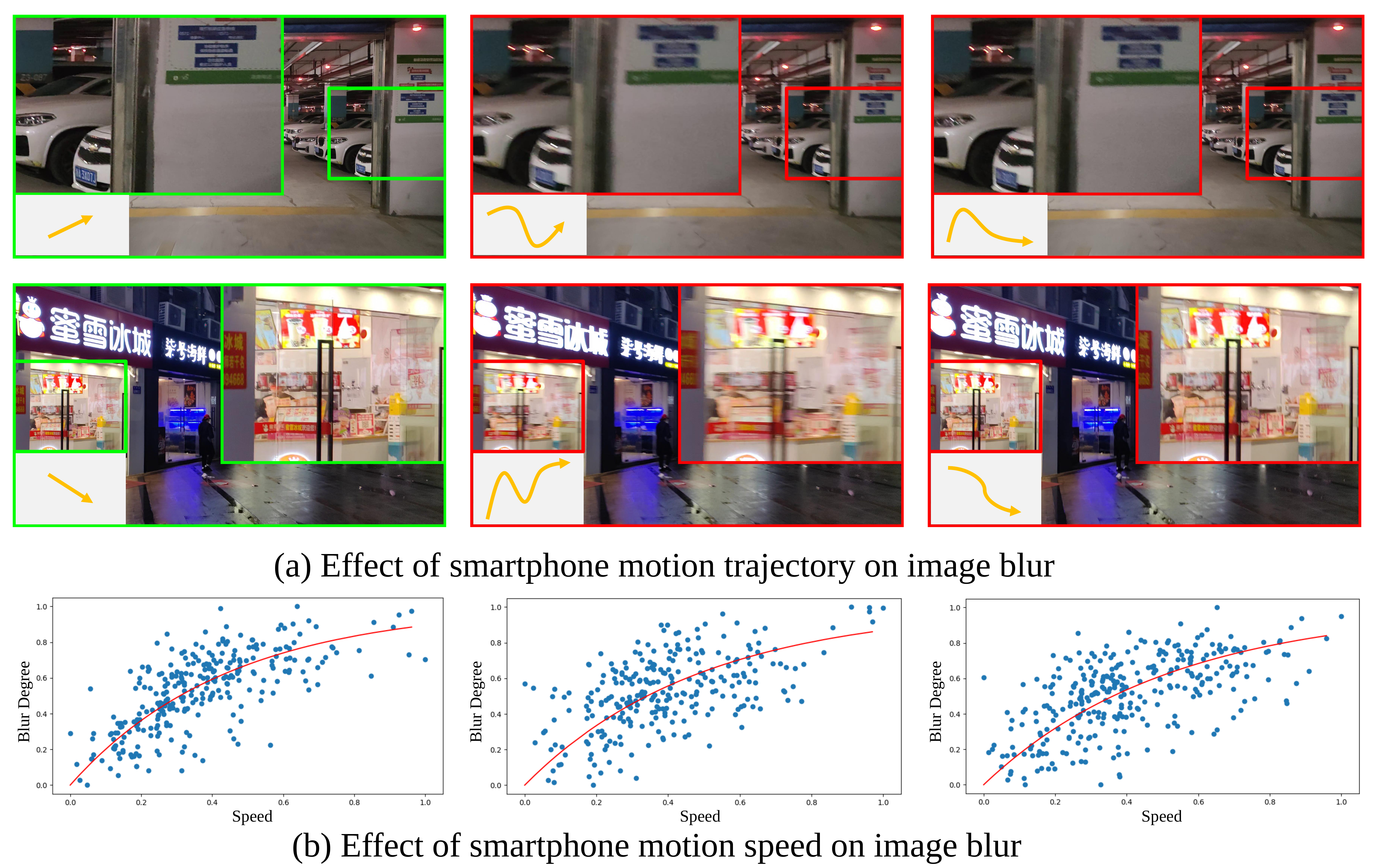}
    \end{subfigure}
    \caption{(a) Effect of the shooting device (with OIS) motion trajectory on image blur. The bottom left corner of the image shows a rough trajectory representation during the exposure time. The more complex the trajectory, the higher the probability of blurry frames (red box). Sharp frames (green box) can be captured when the trajectory is simple and even straight. (b) Effect of the shooting device (with OIS) motion speed on image blur. We randomly select 3 videos and count the correlation between the blur degree and the motion speed.  The faster the speed, the higher the probability of blurring.}
    \label{fig:combined}
\end{figure}

Videos captured by handheld shooting devices generally suffer from significant blur due to shaking caused by hands, vibration caused by walking, and other instability factors. 
Modern shooting devices, such as smartphones, are often equipped with image stabilization technologies to alleviate this problem. 
Take commonly used Optical Image Stabilization (OIS) as an example, it has become a standard feature integrated by smartphone manufacturers, \eg, Huawei, Xiaomi and Apple, which uses a Micro-Electro-Mechanical System (MEMS) gyroscope and gravity sensor to detect movement and adjust the imaging system. For example, if the shooting device is moved slightly to the left, the OIS would perceive this, and then move the lens module or imaging sensor slightly to the right accordingly. However, when the shooting device motion trajectory is complex or its motion speed is fast, the OIS may fail due to its own ability limitations and untimely response. In this case, blur may still appear in the video (see Fig.~\ref{fig:combined}).

A mechanical solution is to use an additional gimbal stabilizer to clamp the shooting device to shoot. But it requires extra money costs and carrying space. Another solution is to post-process the blurry video, \ie, video deblurring. It is relatively cheap and can be integrated into the shooting device system. Recently, learning-based image~\cite{dong2023multi,tsai2022stripformer,zhang2020deblurring,zamir2021multi,kupyn2019deblurgan, SelfIR, lin2024unirestorer} and video deblurring methods~\cite{zhong2020efficient, chan2022basicvsrpp, zhang2022spatio, fgst, Pan_2023_CVPR, Li_2023_CVPR, liang2022vrt, liang2022rvrt, nah2019recurrent, zhang2024bsstnet, chen2024unsupervised} have made great progress, especially in network design. Nevertheless, their pre-trained models are usually effective only on blurry data that is similar to the training samples, and their generalization ability is worrying. Specifically on the handheld video, the image blur is not only affected by camera shake, but also by OIS correction, so its blur distribution is significantly different from that in the existing training datasets (\eg, GoPro~\cite{Nah_2017_CVPR}, BSD~\cite{zhong2023real}). It leads the existing models to perform poorly on handheld video deblurring. 

To address the issue, a straightforward idea is to collect paired handheld video deblurring datasets to train existing networks. But it is with high cost and the process of capturing and pre-processing datasets can be quite complex and cumbersome.
Fortunately, when the shooting device motion trajectory is simple (\eg, a straight line) and the speed is slow, OIS may work well, so that sharp frames can be obtained, as shown in Fig.~\ref{fig:combined}. Sharp frames have the potential to provide deblurring clues and supervision for neighboring blurry frames. Thus, it may be feasible to learn a video deblurring model in a self-supervised manner, which can circumvent the need for paired data. 

Specifically, in this work, we propose a self-supervised handheld video deblurring method, named SelfHVD.
First, we divide the video into multiple segments and select at least one sharpest frame from each segment. Then we align the frame with the neighboring blurry ones, thus the aligned sharp frame can be taken as supervision of the video deblurring model. 
Second, using the above strategy is not enough, as the upper limit of the deblurring model trained in this way is only the selected sharp frame that may be suboptimally clear or not cover sufficient blurry areas. To further improve the deblurring model, we propose a novel Self-Enhanced Video Deblurring (SEVD) method, which utilizes deblurring ability of the previously trained model to construct higher-quality paired training data. In the paired data, SEVD takes the blurry video with some sharp clues removed as the input, and uses the deblurring results (generated from the blurry video with sharp clues) or the selected sharp frame as supervision. In this way, the model can learn to produce results that surpass the sharpest frame in the input segment.
Third, as the number of training iterations increases, the misalignment between supervision and input leads to the spatial position shift between output and input. To avoid this issue, we propose a Self-Constrained Spatial Consistency Maintenance (SCSCM) method. SCSCM constrains the current deblurring results from being spatially consistent with the results generated by the earlier deblurring model, as we observe that early deblurring models generally do not yet suffer from the position shift issue. Moreover, SCSCM can help optimize the model more steadily as training progresses.

For method validation and evaluation, we construct a synthetic dataset (GoProShake) and a real-world dataset (HVD) by collecting hundreds of real-world handheld videos with HUAWEI P40. Extensive experiments on these two and other common real-world datasets. 
We note some self-supervised deblurring methods~\cite{ren_deblur, He2024DADeblur} designed for general blurry video. They usually train the deblurring model by constructing paired data from blurry video, where the sharp information from the video is taken as supervision and is blurred as input. However, their synthetic blurry videos still differ from real-world blurry ones, which prevents their models from performing satisfactorily. From the experimental results, our SelfHVD achieves a significant improvement over them.

The contributions are summarized as follows.
\begin{itemize}[leftmargin=12pt, itemsep=0pt, topsep=0pt]
    \item Based on the observation that sharp clues exist in handheld blurry video, we explore a self-supervised method for handheld video deblurring.
    \item We propose a novel Self-Enhanced Video Deblurring (SEVD) method to improve the deblurring ability of the model and a Self-Constrained Spatial Consistency Maintenance (SCSCM) method to regularize the model to prevent position shift between the output and input.
    \item We construct a synthetic dataset (GoProShake) and a real-world dataset (HVD) for handheld video deblurring. Extensive experiments on these two and other common real-world datasets demonstrate that our method significantly outperforms existing self-supervised ones.
\end{itemize}

%% file: sec/2_relatedwork.tex
\section{Related Work}
\label{sec:relatedwork}
\subsection{Supervised Image Deblurring}
Traditional image deblurring techniques often utilize variational optimization~\cite{fergus2006removing, krishnan2011blind, levin2011efficient, michaeli2014blind}, which depend on prior assumptions about blur kernels and images to tackle the ill-posed nature of the inverse problem. With the rise of deep learning, substantial progress has been achieved~\cite{Nah_2017_CVPR, chen2022simple, zamir2022restormer, liang2021swinir, pham2024blur2blur}. Nah~\etal~\cite{Nah_2017_CVPR} proposed a CNN-based model to deblur without blur kernel estimation. Chen~\etal~\cite{chen2022simple} introduced a simple but effective baseline network for image deblurring. Restormer~\cite{zamir2022restormer} utilized a transformer-based architecture to restore images. Furthermore, SwinIR~\cite{liang2021swinir} designed networks based on the Swin Transformer~\cite{liu2021swin}. For processing unknown blur, Blur2Blur~\cite{pham2024blur2blur} proposed to transform a blurry image into another image with known blur, thus being more amenable for deblurring.

\subsection{Supervised Video Deblurring}
Several synthetic~\cite{Nah_2017_CVPR, Nah_2019_CVPR_Workshops_REDS, su2017deep} and real-world~\cite{zhong2023real,rim2020real} datasets have been used to train supervised video deblurring models. Compared to image deblurring, video deblurring can leverage spatio-temporal information within videos to enhance model performance. On the one hand, several methods~\cite{zhong2020efficient, chan2022basicvsrpp, zhang2022spatio, fgst, Pan_2023_CVPR, Li_2023_CVPR} have employed RNN-based models to leverage spatio-temporal information in videos. IFIRNN\cite{zhong2020efficient} iteratively updates the hidden state via reusing RNN cell parameters. ESTRNN~\cite{zhong2020efficient} employs a GSA module to catch spatially and temporally varying blurs. BasicVSR++~\cite{chan2022basicvsrpp} adopts aggressive bidirectional propagation. STDAN~\cite{zhang2022spatio} and FGST~\cite{fgst} utilize flow-guided attention to align and fuse information from adjacent frames. DSTNet~\cite{Pan_2023_CVPR} develops a wavelet-based feature propagation technique to transfer features in the frequency domain. ShiftNet~\cite{Li_2023_CVPR} proposes a grouped spatio-temporal shift operation to aggregate spatio-temporal features efficiently. On the other hand,  several studies~\cite{liang2022vrt, liang2022rvrt, zhang2024bsstnet} have explored Transformer-based architectures for video deblurring. VRT~\cite{liang2022vrt} utilizes a spatio-temporal self-attention mechanism to integrate information across video frames. RVRT~\cite{liang2022rvrt} proposed a recurrent video restoration transformer with guided deformable attention. BSSTNet~\cite{zhang2024bsstnet} converts the originally dense attention into a sparse form, enabling a more extensive utilization of information throughout the entire video sequence. 
In addition, some methods~\cite{shang2025aggregating,tian2025video, shang2021bringing} utilize sharp frames in input videos to improve video deblurring.
Although these supervised video deblurring methods have impressive results on corresponding datasets, they still perform poorly on real-world videos with unseen blur, \eg, handheld blurry ones.

\subsection{Self-Supervised Deblurring}
Paired real-world data is difficult to obtain, thus some methods~\cite{chi2021test, Nah2021CleanIA, ren_deblur, liu2022meta, He2024DADeblur, wu2024deblur4dgs} have proposed to learn video deblurring models in a self-supervision manner. Chi~\etal~\cite{chi2021test} build a self-supervised auxiliary reconstruction task that shares a portion of the network with the primary deblurring task. Motivated that an ideal deblurring result should contain zero-magnitude motion blur that is hard to be amplified, Nah~\etal~\cite{Nah2021CleanIA} proposed a novel reblurring loss to make the result sharper. Ren~\etal~\cite{ren_deblur} suggested blurring sharp frames in the video using randomly generated blur kernels to obtain paired data for training the model. Furthermore, Liu~\etal~\cite{liu2022meta} utilized GAN~\cite{goodfellow2020generative} to optimize a blurring model in an unpaired training manner, and  DaDeblur~\cite{He2024DADeblur} used a diffusion-based blurring model~\cite{wu2024id} to blur sharp images for fine-tuning deblurring model. 
However, artificially blurry images still differ from real blurry ones. 

%% file: sec/3_method.tex
\section{Proposed Method}
\label{sec:method}
\subsection{Selecting Sharp Frames as Supervision}
\label{subsec:3.1}

\noindent\textbf{Characteristic of Handheld Video.} Image stabilization technologies, such as Electronic Image Stabilization (EIS) and Optical Image Stabilization (OIS) have been commonly used in modern shooting devices. They first obtain the motion information from the Inertial Measurement Unit (IMU) (including gyroscope and gravity sensor), then perform attitude calculation~\cite{li2013extended, madgwick2010efficient, sabatelli2012double, bell2014non} on the IMU data. During the calculation, Kalman filter~\cite{kalman1960new}, Mahony filter~\cite{mahony2008nonlinear}, and others are used to smooth the current data according to historical data.
Finally, motion compensation is carried out according to acquired motion information to achieve image stabilization. 
Such stabilization technologies usually work well when the shooting device motion trajectory is simple and the speed is slow, but they may fail when the trajectory is complex or the speed is fast. Therefore, blurry frames and sharp frames often coexist in handheld videos, which gives us an opportunity to explore self-supervised methods for handheld video deblurring, with further discussion in Sec.~\textcolor{red}{A}
of the supplementary material.

\noindent\textbf{Selecting Sharp Frames.} A straightforward idea is to select these sharp frames as supervision of the video deblurring model.
As the sharp frame detection approach suggested by Ren~\etal~\cite{ren_deblur}, the variance of the image Laplacian can be considered as a measurement of sharpness degree. Given an image $\mathbf{I}$, the variance of its Laplacian is:
\begin{equation}
    v_l(\mathbf{I}) = \mathbb{E}[(\Delta \mathbf{I} -\overline{\Delta \mathbf{I}})^{2}],
\label{equ:3.1}
\end{equation}
where $\Delta \mathbf{I}$ is the image Laplacian obtained by convolving $\mathbf{I}$ with the Laplacian mask, and $\overline{\Delta \mathbf{I}}$ is the mean value of $\Delta \mathbf{I}$. 
Then, we construct a histogram of the sharpness degrees, and expect the histogram to exhibit a bimodal distribution, with the two peaks corresponding to sharp and blurry frames, respectively. To define the threshold that separates the two classes, we use an automatic image thresholding technique, \ie, Otsu's method~\cite{otsu1975threshold}, which minimizes the intra-class variance based on the histogram of sharpness levels. 
Frames with sharpness values below the threshold are classified as blurry ones, while those above the threshold are classified as sharp frames. 

When applying the above selection method to the whole video, the selected sharp frames may be unevenly distributed, resulting in failure to cover most of the video scenes. Thus, we further split the video into segments, and for a video segment without the sharp frames, we regard the frame with the highest $v_l$ as its sharp frame. 
Finally, we define these sharp frames determined by the above global and local selection steps as $\mathbf S$. In our implementation, we divide the video into segments of 20 frames, and the accuracy is computed by comparing our selected sharp frames with manually labeled ones, reaching 96.77\% on GoProShake and 91.88\% on HVD. \\
\begin{figure*}
  \centering
  \includegraphics[width=0.99\linewidth]{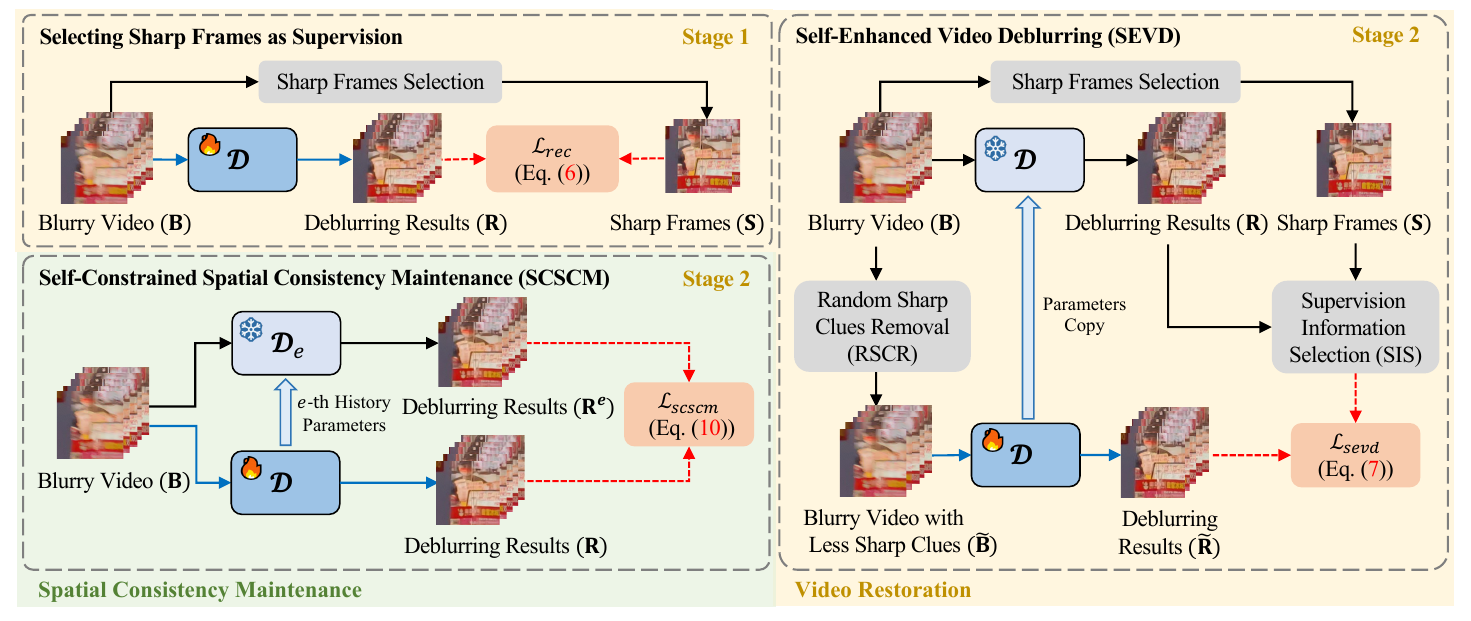}
  \caption{Overview of our SelfHVD. Given a blurry video captured by a handheld shooting device, we first select the sharp frames and take them as misalignment labels. Then, Self-Enhanced Video Deblurring (SEVD) constructs higher-quality paired training data to further improve the model performance. Self-Constrained Spatial Consistency Maintenance (SCSCM) is proposed to prevent position shifts between output and input frames.}
  \label{fig:3.2}
  \vspace{-4mm}
\end{figure*}

\noindent\textbf{Taking Sharp Frames as Supervision.} Given a handheld blurry video $\mathbf B$ consisting of $N$ frames ${\mathbf B_n}_{n=1}^N$, video deblurring aims to restore the corresponding sharp components ${{\mathbf R}_n}_{n=1}^N$.
When taking the selected sharp frames $\mathbf S$ as supervision, the optimal parameters $\mathbf{\Theta}_\mathcal{D}$ of deblurring model $\mathcal{D}$ can be formulated as,
\begin{equation}
    \mathbf \Theta_{\mathcal{D}}^*=\arg\min_{\mathbf{\Theta}_{\mathcal{D}}}\mathcal{L}\left(\mathcal{D}\left(\mathbf B; \mathbf \Theta_{\mathcal{D}}\right), \mathbf S\right),
\label{equ:3.3}
\end{equation}
where $\mathcal{L}$ denotes the loss function.

Specifically, for the input frame $\mathbf B_i$, we identify the temporally closest sharp frame $\mathbf S_j$ as its misalignment label, where $j = \mathcal{J}(i)$ and $\mathcal{J}$ denotes the function that determines the temporally closest frame index. We use optical flow model (\ie, SEA-RAFT~\cite{wang2024sea}) to align output ${\mathbf R}_i$ and label $\mathbf S_j$.
The optical flow from ${\mathbf R}_i$ to $\mathbf S_j$ can be denoted as $\mathbf \Phi_{i\rightarrow j}$.
Then, the sharp frame $\mathbf S_j$ can be backward warped via bilinear resampling, \ie,
\begin{equation}
    \mathbf S_{j\rightarrow i} = \mathcal{W}(\mathbf S_j, \mathbf \Phi_{i\rightarrow j}),
\label{equ:3.9}
\end{equation}
where $\mathcal{W}$ is the warping operation.
Moreover, we design two masks to exclude incorrect alignment and occluded regions, respectively.
The former mask can be estimated by the uncertainty map of the optical flow, and SEA-RAFT~\cite{wang2024sea} can directly output it. Thus, the mask can be written as,
\begin{equation}
    \mathbf M_{uncer}^i = \mathbf U_{j\rightarrow i} \odot \mathcal{W}(\mathbf U_{i\rightarrow j}, \mathbf \Phi_{i\rightarrow j}),
   \label{equ:3.11}
\end{equation}
where $\odot$ is the pixel-wise multiplication operation. $\mathbf U_{j\rightarrow i}$ and 
$\mathbf U_{i\rightarrow j}$ represents the  uncertainty map of $\mathbf \Phi_{j\rightarrow i}$ and $\mathbf \Phi_{i\rightarrow j}$, respectively. 
The latter mask can be estimated using the forward-backward flow consistency~\cite{alvarez2007symmetrical}, and it can be written as,
\begin{equation}
   \mathbf M_{occ}^i=\min(s||\mathcal{W}(\mathcal{W}(\mathbf G; \mathbf \Phi_{j\rightarrow i}); \mathbf \Phi_{i\rightarrow j})-\mathbf G||_{2}, \mathbf{1}),
\label{equ:3.12}
\end{equation}
where $\mathbf G$ is an image coordinate map. The scaling factor $s$ controls the strength of the occlusion map. 
Finally, the reconstruction loss $\mathcal{L}_{rec}$ can be formulated as,
\begin{equation}
    \mathcal{L}_{rec} = \frac{1}{N} \sum_{i=1}^N||\mathbf M_i \odot({\mathbf R}_i - \mathbf S_{j\rightarrow i})||_1,
\label{equ:3.14}
\end{equation}
where ${\mathbf R}_i=\mathcal{D}(\mathbf B; \mathbf \Theta_{\mathcal{D}})_i$ and $\mathbf M_i= \mathbf M_{uncer}^i \odot \mathbf M_{occ}^i$, as validated through ablation studies in Sec.~\textcolor{red}{D.1}
and further shown in Fig.~\textcolor{red}{J}
and Fig.~\textcolor{red}{K}
in the supplementary material.

\subsection{Self-Enhanced Video Deblurring}
\label{subsec:3.2}
As previously described, we select the relatively sharp frames from each video segment as supervision of the deblurring model. However, it may not be enough, as the upper limit of the deblurring model trained in this way is only the selected sharp frame, while the sharp frame may be suboptimally clear, and the aligned sharp frame may not cover sufficient blurry areas. To address this issue, we propose the Self-Enhanced Video Deblurring (SEVD) method to utilize the existing deblurring ability of the model to construct higher-quality paired training data. This not only improves the overall deblurring performance but also enables the model to handle object motion blur.

First, we use a Random Sharp Clues Removal (RSCR)  strategy to randomly remove the sharp clues from the input video and replace them with adjacent blurry frames. Specifically, we use the method proposed in \textbf{Selecting Sharp Frames} (Sec.~\ref{subsec:3.1}) to distinguish between blurry and sharp frames in every video segment. Define the number of sharp frames as $L$, then we randomly replace $l~(0 < l \le L)$ sharp frames with the temporally closest blurry frames to obtain the video $\tilde{\mathbf B}$, which has less sharp clues. 
Thus, the sharp frame $\mathbf S$ is with higher quality than the clearest frame in $\tilde{\mathbf B}$. Taking  $\mathbf S$  as supervision of input video $\tilde{\mathbf B}$ can help model to break through its own input (\ie, $\tilde{\mathbf B}$) to learn better results.

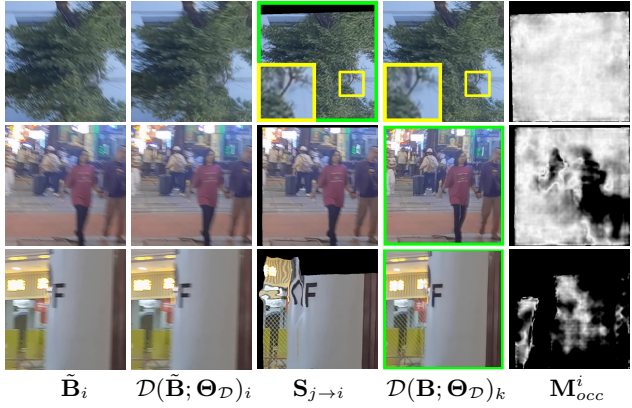
\begin{figure}[t]
    \centering
    \begin{overpic}[percent,width=0.99\linewidth]{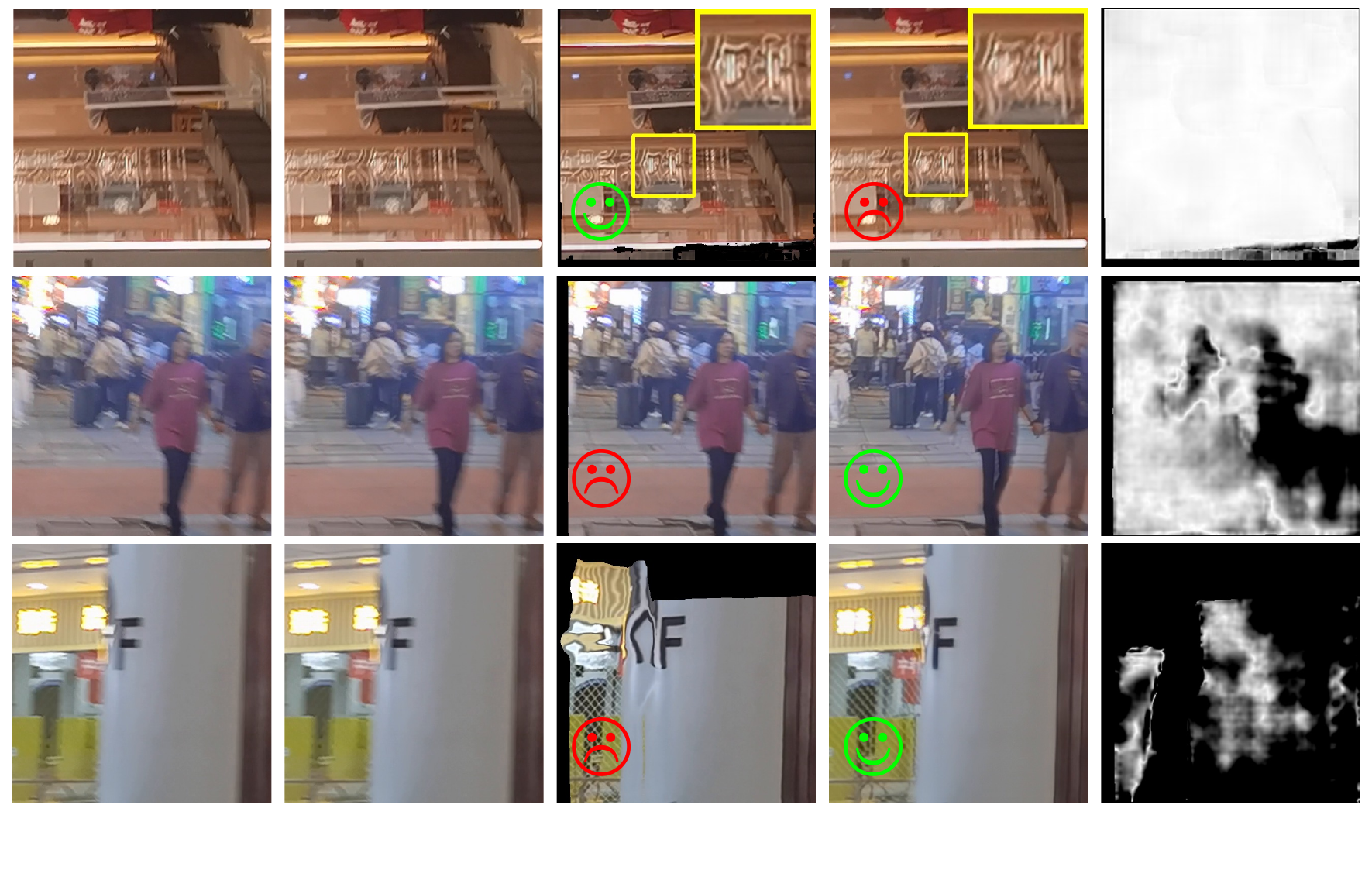}
    \put(9,0.3){\footnotesize{$\tilde{\mathbf{B}}_i$}}
    \put(21,0.3){\footnotesize{$\mathcal{D}(\tilde{\mathbf{B}}; \mathbf{\Theta}_{\mathcal{D}})_i$}}
    \put(46,0.3){\footnotesize{$\mathbf{S}_{j\rightarrow i}$}}
    \put(61,0.3){\footnotesize{$\mathcal{D}(\mathbf{B}; \mathbf{\Theta}_{\mathcal{D}})_k$}}
    \put(87,0.3){\footnotesize{$\mathbf{M}_{occ}^i$}}
    \end{overpic}
    \caption{From left to right: sharp-clues-less blurry video, deblurring result of sharp-clues-less blurry video, warped sharp frame, deblurring result of original input video, occlusion map. From top to bottom, the unmasked region ratio in $\mathbf{M}_{occ}^i$ is $0.90$, $0.64$, and $0.16$. The smiley denotes the final supervision for $\mathcal{D}(\tilde{\mathbf{B}}; \mathbf{\Theta}_{\mathcal{D}})_i$.}
    \label{fig:sevd}
\end{figure}

Second, since the selected sharp frame may not cover sufficient blurry areas and some aligned areas will be excluded (see Eq.~(\ref{equ:3.14})) due to inaccurate optical flow and occlusion, only using sharp frames as supervision may be not effective enough. In fact, the result $\mathcal{D}(\mathbf B; \mathbf \Theta_{\mathcal{D}})$ (\ie, deblurring the original input video $\mathbf B$) may be not less clear than $\mathbf S$. More importantly, $\mathcal{D}(\tilde{\mathbf B}; \mathbf \Theta_{\mathcal{D}})$ is perfectly aligned with the corresponding frame in $\mathcal{D}(\mathbf B; \mathbf \Theta_{\mathcal{D}})$. Thus, we can further take $\mathcal{D}(\mathbf B; \mathbf \Theta_{\mathcal{D}})$ as the target of $\mathcal{D}(\tilde{\mathbf B}; \mathbf \Theta_{\mathcal{D}})$.

In practice, in order to use higher-quality images as supervision of input $\tilde{\mathbf B}$, we suggest a Supervision Information Selection (SIS) strategy to select a better one from $\mathbf S$ and $\mathcal{D}(\mathbf B; \mathbf \Theta_{\mathcal{D}})$.
Specifically, if the aligned sharp frame $\mathbf{S}_{j\rightarrow i}$ does not exhibit excessive distortion due to content difference between $\mathbf{S}_{j}$ and $\mathcal{D}(\tilde{\mathbf B}; \mathbf \Theta_{\mathcal{D}})_i$, and is sharper than the corresponding $\mathcal{D}(\mathbf B; \mathbf \Theta_{\mathcal{D}})$, as shown in the top row of Fig.~\ref{fig:sevd}, we take $\mathbf S_{j\rightarrow i}$ as the supervision. It is noted that we consider distortion to have occurred when the proportion of unmasked regions in the occlusion map $\mathbf M_{occ}$ is below a distortion threshold. Otherwise, as shown in the middle and bottom row of Fig.~\ref{fig:sevd}, $\mathcal{D}(\mathbf B; \mathbf \Theta_{\mathcal{D}})_k$ is better than the aligned sharp frame, thus we choose $\mathcal{D}(\mathbf B; \mathbf \Theta_{\mathcal{D}})_k$ to serve as supervision. Finally, $\mathcal{L}_{\text{sevd}}$ can be defined as
\begin{equation}
    \mathcal{L}_{sevd}  \! = \! \frac{1}{N} \! \sum_{i=1}^N \! 
    \begin{cases}{}
         \! \left\| \mathbf{M}_i \! \odot \! \left(\tilde{\mathbf R}_i \! - \! \mathbf{S}_{j \rightarrow i} \right) \right\|_1 \!\!\!\!\!  & \text{if $c$ is True},    \\[2mm]
         \left\| \tilde{\mathbf R}_i \! - \! sg\left({\mathbf R}_k\right) \right\|_1 \!\!\!\!\!   & \text{if $c$ is False},    \\
    \end{cases}
\label{equ:sis_split}
\end{equation}
\begin{equation}
    \text{$c = \mathrm{mean}(\mathbf{M}_{occ}^i) > \tau$ and  $v_l(\mathbf{S}_{j \rightarrow i}) > v_l(\mathbf{R}_k)$},
\label{equ:sis_split_cond}
\end{equation}
where $\tilde{\mathbf R}_i=\mathcal{D}(\tilde{\mathbf{B}}; \boldsymbol{\Theta}_{\mathcal{D}})_i$, ${\mathbf R}_k=\mathcal{D}(\mathbf{B}; \boldsymbol{\Theta}_{\mathcal{D}})_k$, $k$ is the frame index of ${\mathbf R}$ corresponding to $i$-th frame in $\tilde{\mathbf R}$, $\mathrm{mean}(\mathbf{M}_{occ}^i)$ is the unmasked regions proportion, $\tau$ is a threshold, $v_l(\cdot)$ represents sharpness (see Eq.~(\ref{equ:3.1})), and $sg(\cdot)$ is the stop gradient operation. 

SEVD effectively enhances the performance in removing camera motion blur. Beyond that, it also enables the restoration of object motion blur. Specifically, object motion is typically non-uniform, so relatively sharp content is retained when the object is still or moves slowly. Video deblurring models can aggregate information across multiple frames, allowing these sharp contents to provide crucial clues for dealing with blur when the object moves fast. Compared to deblurring results from the blurry video without sharp clues, those from the original blurry video perform better on object motion blur, as illustrated in the middle row of Fig.~\ref{fig:sevd} and Fig.~\textcolor{red}{E}
in the supplementary material. As a result, the high-quality training pairs constructed by SEVD also offer more reliable supervision for object motion deblurring.

\begin{figure}[t]
    \centering
    \includegraphics[width=0.99\linewidth]{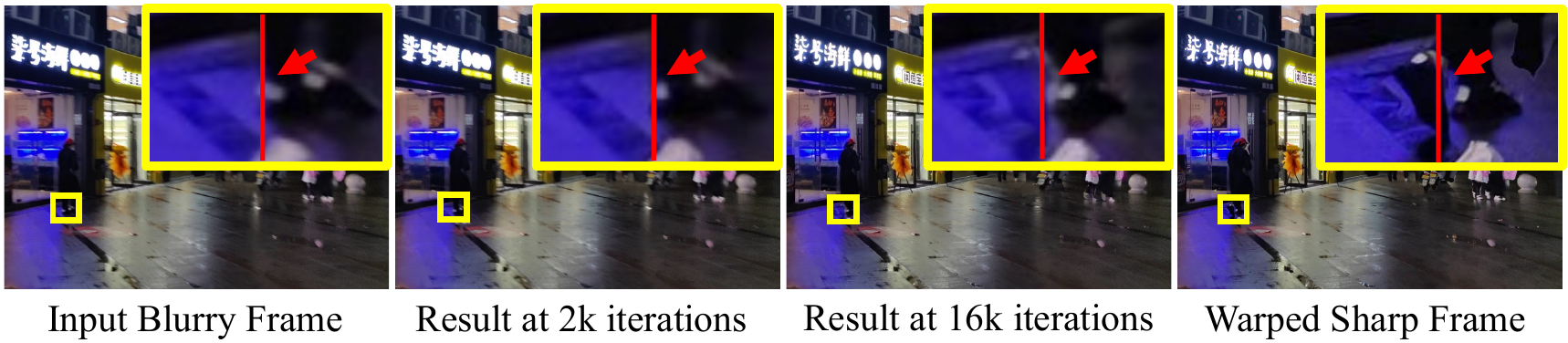}
    \caption{Vanilla self-supervised training performs well in early stages, but fails to maintain spatial consistency between input and output as training progresses.}
    \label{fig:3.6}
\end{figure}

\subsection{Self-Constrained Spatial Consistency Maintenance}
\label{subsec:3.3}
Although we have carefully designed the self-supervised deblurring method, the spatial inconsistency issue between output and input is prone to occur as training progresses, as shown in Fig.~\ref{fig:3.6} (c). In fact, it is hard to make sure of perfect alignment between the aligned sharp frames and the input frames, even using the most advanced optical flow network. Once alignment errors occur, even small ones, the model can gradually learn to shift the position of the input, bringing adverse effects on subsequent model training.

Fortunately, information bottleneck theory~\cite{tishby2015deep} has indicated that mutual information between network characteristics and the input increases first and then decreases during training~\cite{liu2024learning}. As illustrated in Fig.~\ref{fig:3.6}, in the early stage, spatial consistency can be maintained; in the later stage, the inconsistency issue starts to occur. Inspired by this, we propose Self-Constrained Spatial Consistency Maintenance (SCSCM), using the historical result as auxiliary supervision to help the deblurring model maintain spatial consistency between the input and output. 

Specifically, we denote the historical model parameters at $e$-th training iterations as $\mathbf{\Theta}_{\mathcal{D}_e}$. The output ${\mathbf{R}}^e$ from  $\mathbf{\Theta}_{\mathcal{D}_e}$ is both sharper than the input $\mathbf B$ and aligned with  , and it can be written as,
\begin{equation}
    {\mathbf{R}}^e=\mathcal{D}(\mathbf B; \mathbf \mathbf{\Theta}_{\mathcal{D}_e}).
\label{equ:g}
\end{equation}
To ensure that the model parameters are updated under the supervision of sharper frames while remaining close to previously learned ones that implicitly preserve spatial alignment with the input, we incorporate SCSCM into the learning objective after $e$ training iterations, defined as:
\begin{equation}
     \mathcal{L}_{scscm} = \frac{1}{N} \sum_{i=1}^N||\tilde{\mathbf R}_i - sg({\mathbf{R}}_k^e)||_1,
\label{equ:3.20}
\end{equation}
where $e$ would be updated with the latest training \#iterations if the output from the latest model is still aligned with the input and is sharper than the previous output.

\begin{table*}[t]
\centering
\caption{Quantitative comparison on the synthetic GoProShake and real-world HVD datasets. `$Network$' in `SelfHVD$_{Network}$'
denotes the deblurring network we use, where ESTRNN~\cite{zhong2020efficient} is also adopted by Ren~\etal~\cite{ren_deblur} and DaDeblur~\cite{He2024DADeblur}.}
\vspace{-2mm}
\resizebox{1.\linewidth}{!}{
\begin{tabular}{clcccc}
\hline
\multicolumn{2}{c}{\multirow{2}{*}{Methods}}                                                                 & GoProShake                                                           & HVD-Huawei                                                           & HVD-Xiaomi                                                           & HVD-iPhone                                                           \\
\multicolumn{2}{c}{}                                                                                         & PSNR$\uparrow$ / SSIM$\uparrow$                                     & MUSIQ$\uparrow$ / MANIQA$\uparrow$                                   & MUSIQ$\uparrow$ / MANIQA$\uparrow$                                   & MUSIQ$\uparrow$ / MANIQA$\uparrow$                                   \\ \hline
\multirow{4}{*}{\begin{tabular}[c]{@{}c@{}}Fully-\\ Supervised\end{tabular}} 
& IFIRNN~\cite{nah2019recurrent}      & 34.66 / 0.9448                           & 24.1043 / 0.1916                           & 29.7281 / 0.2212                           & 22.3710 / 0.2535                           \\
& ESTRNN~\cite{zhong2020efficient}    & 34.19 / 0.9369                           & 24.0383 / 0.1917                           & 29.4117 / 0.2193                           & 21.6487 / 0.2506                           \\
& RVRT~\cite{liang2022rvrt}           & 37.02 / 0.9473                           & 24.9269 / 0.1923                                      & \textbf{30.2976} / 0.2215                           & 22.4535 / 0.2546                           \\
& BasicVSR++~\cite{chan2022basicvsrpp}& \textbf{37.99} / \textbf{0.9683}         & \textbf{25.2499} / \textbf{0.2006}         & 30.0775 / \textbf{0.2235}                & \textbf{22.6709} / \textbf{0.2564}       \\ \hline
\multirow{6}{*}{\begin{tabular}[c]{@{}c@{}}Self-\\ Supervised\end{tabular}}  
& Ren~\etal~\cite{ren_deblur}         & 25.05 / 0.7428                           & 22.5433 / 0.1771                           & 22.7757 / 0.2193                           & 20.2299 / 0.2653                           \\
& DaDeblur~\cite{He2024DADeblur}      & 29.54 / 0.8772                           & 26.8422 / 0.2025                           & 32.3833 / 0.2322       & 25.3244 / 0.2684                \\
& SelfHVD$_{IFIRNN}$                  & 34.32 / 0.9302                           & 27.6922 / 0.2137                                      & 32.7765 / 0.2530                           & 25.7711 / 0.2756                           \\
& SelfHVD$_{ESTRNN}$                  & 33.60 / 0.9216                           & 27.6873 / 0.2126                                      & 32.4002 / \textbf{0.2574}                           & 25.7556 / \textbf{0.2842}                           \\
& SelfHVD$_{RVRT}$                    & 36.31 / 0.9300                           & 27.8345 / 0.2088                                      & 32.4416 / 0.2303                           & \textbf{25.8437} / 0.2711                           \\
& SelfHVD$_{BasicVSR++}$              & \textbf{37.44} / \textbf{0.9359}         & \textbf{28.0040} / \textbf{0.2175}         & \textbf{32.8564} / 0.2236       & 25.7022 / 0.2686       \\ \hline
\end{tabular}
}
\label{tab:5.2}
\end{table*}

\begin{table*}[t]
\centering
\setlength{\tabcolsep}{1.8mm}
\caption{
Quantitative results on BSD~\cite{zhong2020efficient}, RealBlur~\cite{rim2020real} and RBVD~\cite{chao2022} via test-time fine-tuning. Each group reports PSNR and SSIM. All methods use ESTRNN~\cite{zhong2020efficient} as the deblurring network, with its GoPro-trained version serving as the baseline.
}
\vspace{-2mm}
\begin{tabular}{lclclclclcl}
\hline
\multicolumn{1}{c}{\multirow{2}{*}{Methods}}              & \multicolumn{2}{c}{BSD-1ms8ms~\cite{zhong2020efficient}} & \multicolumn{2}{c}{BSD-2ms16ms~\cite{zhong2020efficient}} & \multicolumn{2}{c}{BSD-3ms24ms~\cite{zhong2020efficient}} & \multicolumn{2}{c}{RealBlur~\cite{rim2020real}} & \multicolumn{2}{c}{RBVD~\cite{chao2022}} \\
\multicolumn{1}{c}{}                                      & \multicolumn{2}{c}{PSNR$\uparrow$ / SSIM$\uparrow$}         & \multicolumn{2}{c}{PSNR$\uparrow$ / SSIM$\uparrow$}          & \multicolumn{2}{c}{PSNR$\uparrow$ / SSIM$\uparrow$}          & \multicolumn{2}{c}{PSNR$\uparrow$ / SSIM$\uparrow$} & \multicolumn{2}{c}{PSNR$\uparrow$ / SSIM$\uparrow$} \\ \hline
Baseline                                                  & \multicolumn{2}{c}{25.57 / 0.747}                          & \multicolumn{2}{c}{24.64 / 0.726}                           & \multicolumn{2}{c}{26.01 / 0.748}                           & \multicolumn{2}{c}{25.87 / 0.773}                    & \multicolumn{2}{c}{24.47 / 0.725}                 \\
+Blur2Blur~\cite{pham2024blur2blur}                       & \multicolumn{2}{c}{25.64 / 0.750}                          & \multicolumn{2}{c}{24.72 / 0.728}                           & \multicolumn{2}{c}{26.01 / 0.749}                           & \multicolumn{2}{c}{26.16 / 0.800}                    & \multicolumn{2}{c}{24.45 / 0.725}                 \\
+DaDeblur~\cite{He2024DADeblur}                           & \multicolumn{2}{c}{29.44 / 0.843}                          & \multicolumn{2}{c}{28.36 / 0.820}                           & \multicolumn{2}{c}{28.23 / 0.808}                           & \multicolumn{2}{c}{27.41 / 0.819}                    & \multicolumn{2}{c}{27.02 / 0.771}                 \\
+Ours                                                     & \multicolumn{2}{c}{\textbf{31.01 / 0.873}}                 & \multicolumn{2}{c}{\textbf{29.00 / 0.832}}                  & \multicolumn{2}{c}{\textbf{29.31 / 0.825}}                  & \multicolumn{2}{c}{\textbf{28.76 / 0.848}}           & \multicolumn{2}{c}{\textbf{27.69 / 0.785}}        \\ \hline
\end{tabular}
\label{tab:benchmark}
\end{table*}

\begin{figure*}[t!]
    \centering
    \includegraphics[width=0.99\linewidth]{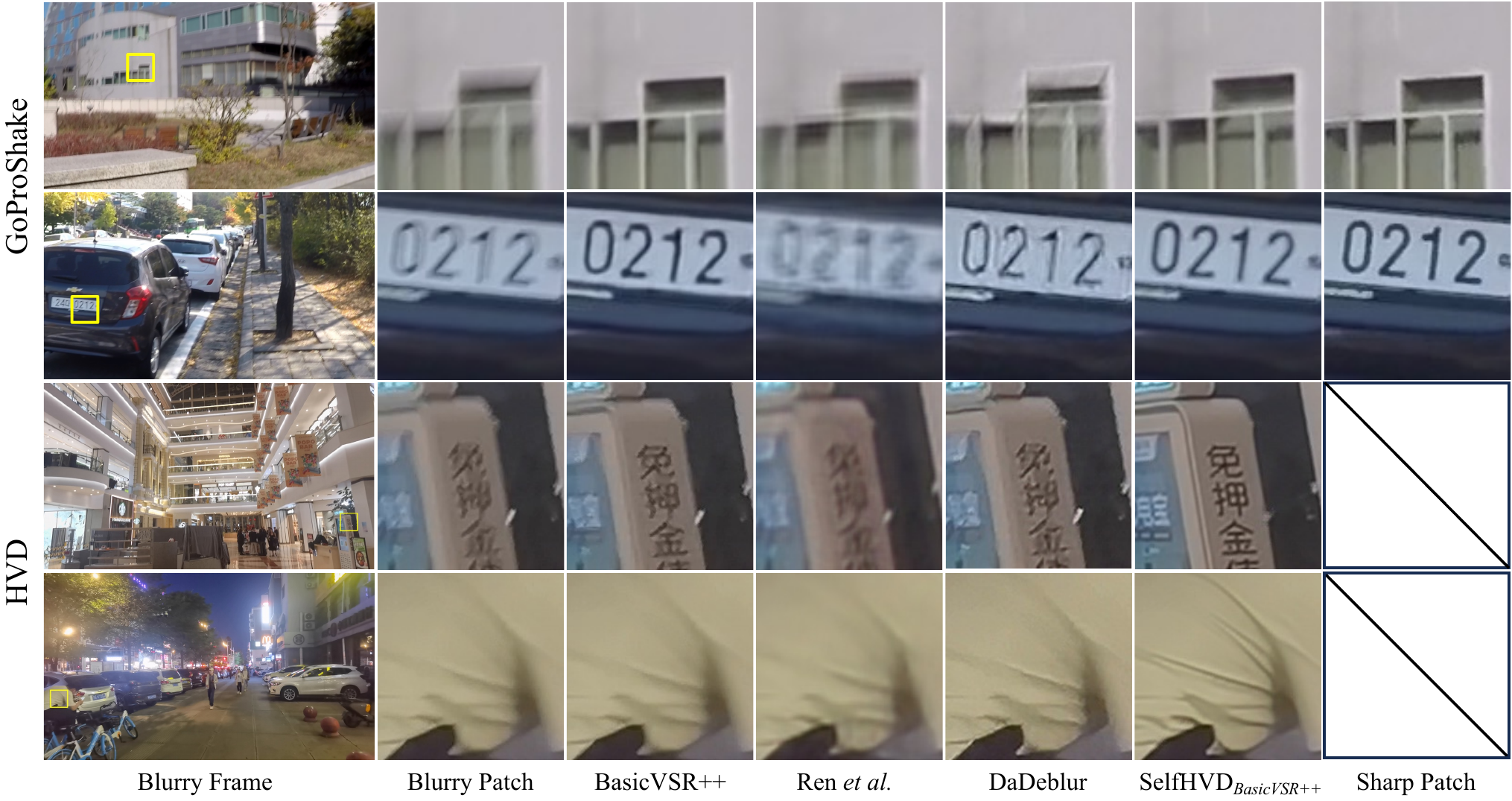}
    \caption{ Qualitative comparison on synthetic GoProShake and real-world HVD datasets.}
    \label{fig:qualitative}
    \vspace{-4mm}
\end{figure*}

\subsection{Learning Objectives}
\label{subsec:3.4}
To further improve the visual quality, we adopt the perceptual loss during training, which can be written as:
\begin{equation}
    \mathcal{L}_{vgg_1} = \frac{1}{N} \sum_{i=1}^N||\mathbf M_i\odot(\phi({\mathbf R}_i) - \phi(\mathbf S_{j\rightarrow i}))||_1,
\label{equ:3.21}
\end{equation}
\begin{equation}
    \mathcal{L}_{vgg_2}  \! = \! \frac{1}{N} \!\!  \sum_{i=1}^N \!\! 
    \begin{cases}{}
         \! \left\| \mathbf{M}_i \! \odot \! \left(\phi(\tilde{\mathbf R}_i) \! - \! \phi(\mathbf{S}_{j \rightarrow i}) \right) \right\|_1   & \!\!\!\! \text{if $c$ is True},    \\[2mm]
         \left\| \phi(\tilde{\mathbf R}_i) \! - \! sg\left(\phi({\mathbf R}_k)\right) \right\|_1    & \!\!\!\! \text{if $c$ is False},    \\
    \end{cases}
\label{equ:3.21_2}
\end{equation}
where $\phi$ is the pre-trained VGG~\cite{simonyan2014very} and $c$ is the select condition (see Eq.~(\ref{equ:sis_split_cond})).

The overall training of our self-supervised handheld video deblurring model consists of two stages, and the learning objective in the first stage can be written as,
\begin{equation}
     \mathcal{L}_{s_1} = \mathcal{L}_{rec} + \beta\mathcal{L}_{vgg_1},
\label{equ:3.22}
\end{equation}
where  $\beta$ is the weight of $\mathcal{L}_{vgg}$ and is set to 1. In the second stage, the model has had a certain deblurring ability, we introduce Self-Enhanced Video Deblurring (SEVD) to improve the quality of training data and deploy Self-Constrained Spatial Consistency Maintenance (SCSCM) to ensure spatial consistency between input and output. The learning objective can be written as,
\begin{equation}
     \mathcal{L}_{s_2} = \mathcal{L}_{sevd} + \lambda\mathcal{L}_{scscm} + \beta\mathcal{L}_{vgg_2},
\label{equ:3.23}
\end{equation}
where $\lambda$ is the weight of $\mathcal{L}_{scscm}$ and is set to 1.

%% file: sec/4_experiment.tex
\section{Experiments}
\label{sec:exp}
\subsection{Implementation Details}
\textbf{Datasets.} To verify the effectiveness of our method, we propose a synthetic handheld video deblurring dataset GoProShake, and a real-world dataset HVD. GoProShake simulates handheld motion with OIS using camera trajectories from MonST3R~\cite{zhang2024monst3r}, while HVD is collected using HUAWEI P40 (HVD-Huawei), Xiaomi 15 (HVD-Xiaomi) and iPhone 16 (HVD-iPhone). Please refer to Sec.~\textcolor{red}{B}
in the supplementary material for detailed descriptions.\\
\textbf{Framework Details.} Note that this work does not focus on the design of network architectures, and we employ existing ones directly. We adopt CNN-based (\ie IFIRNN~\cite{nah2019recurrent}, ESTRNN~\cite{zhong2020efficient}. BasicVSR++~\cite{chan2022basicvsrpp}) and Transformer-based (\ie RVRT~\cite{liang2022rvrt}) video deblurring models as the reconstruction network. We use the pre-trained SEA-RAFT~\cite{wang2024sea} to estimate optical flow. \\
\textbf{Training Details.} During the training phase, the input frames are randomly cropped into patches with resolutions of 256×256, along with the application of random flipping and rotation. During the testing phase, the resolution of frames remains unchanged. The video deblurring model is optimized using Adam optimizer~\cite{kingma2014adam}, where $\beta_1=0.9$ and $\beta_2=0.999$. The initial learning rate is $1e^{-4}$, gradually decayed to $1e^{-7}$ by the cosine annealing strategy. We set random seed to $0$ and the distortion threshold $\tau$ to $0.5$. All experiments are conducted with PyTorch~\cite{paszke2019pytorch} on a single Nvidia GeForce RTX A6000 GPU. \\
\textbf{Evaluation Configurations.} For the synthetic dataset GoProShake, we use PSNR, SSIM~\cite{wang2004image} as evaluation metrics. For the real-world dataset HVD, because there is no ground truth, we use no-reference image quality assessment MUSIQ~\cite{ke2021musiq} and MANIQA~\cite{yang2022maniqa} as evaluation metrics. Additional evaluation details are provided in Sec.~\textcolor{red}{E}
of the supplementary material.

\subsection{Comparison with State-of-the-Arts}
\textbf{Quantitative Analysis.} Table~\ref{tab:5.2} shows the quantitative comparison on the synthetic dataset GoProShake and the real-world dataset HVD. The results show that SelfHVD outperforms the previous self-supervised methods Ren~\etal~\cite{ren_deblur} and DaDeblur~\cite{He2024DADeblur}. Moreover, SelfHVD achieves results comparable to the corresponding supervised methods on GoProShake. For results of supervised pre-training on the other real-world dataset along with their self-supervised fine-tuning performance on HVD, please refer to Sec.~\textcolor{red}{D.5}
in the supplementary material. To further verify the effectiveness of our method, we conduct additional comparisons with DaDeblur~\cite{He2024DADeblur} on publicly available real-world datasets, including BSD~\cite{zhong2020efficient}, RBVD~\cite{chao2022}, and RealBlur~\cite{rim2020real}. As shown in Table~\ref{tab:benchmark}, under the same test-time training setting on ESTRNN~\cite{zhong2020efficient} as DaDeblur~\cite{He2024DADeblur}, our method consistently outperforms both the baseline (i.e., ESTRNN~\cite{zhong2020efficient} trained on GoPro~\cite{Nah_2017_CVPR}) and DaDeblur~\cite{He2024DADeblur}, with less training time (including data preprocessing). On BSD~\cite{zhong2020efficient}, our method consistently surpasses both the baseline and DaDeblur~\cite{He2024DADeblur} under various exposure settings, demonstrating strong robustness. And on RealBlur~\cite{rim2020real} and RBVD~\cite{chao2022}, our approach also achieves the highest PSNR and SSIM. These experimental results show that our method can achieve effective handheld video deblurring through a self-supervised approach without ground truth.\\
\textbf{Qualitative Analysis.} The visual comparisons on GoProShake and HVD datasets are shown in Fig.~\ref{fig:qualitative}. As illustrated, SelfHVD$_{BasicVSR++}$ achieves better deblurring results compared to previous self-supervised methods, even for object motion blur (Sec.~\textcolor{red}{C}
in the supplementary material). Moreover, the results of SelfHVD$_{BasicVSR++}$ are visually comparable to BasicVSR++~\cite{chan2022basicvsrpp} on GoProShake. And under the same test-time training setting as DaDeblur~\cite{He2024DADeblur}, our method also achieves better visual results than DaDeblur~\cite{He2024DADeblur} on BSD~\cite{zhong2020efficient}, RealBlur~\cite{rim2020real} and RBVD~\cite{chao2022}, as shown in Fig.~\ref{fig:qualitative_online}. 
And additional visual results are in Sec.~\textcolor{red}{E}
of the supplementary material.

\begin{figure*}[t]
  \centering
  \includegraphics[width=0.97\linewidth]{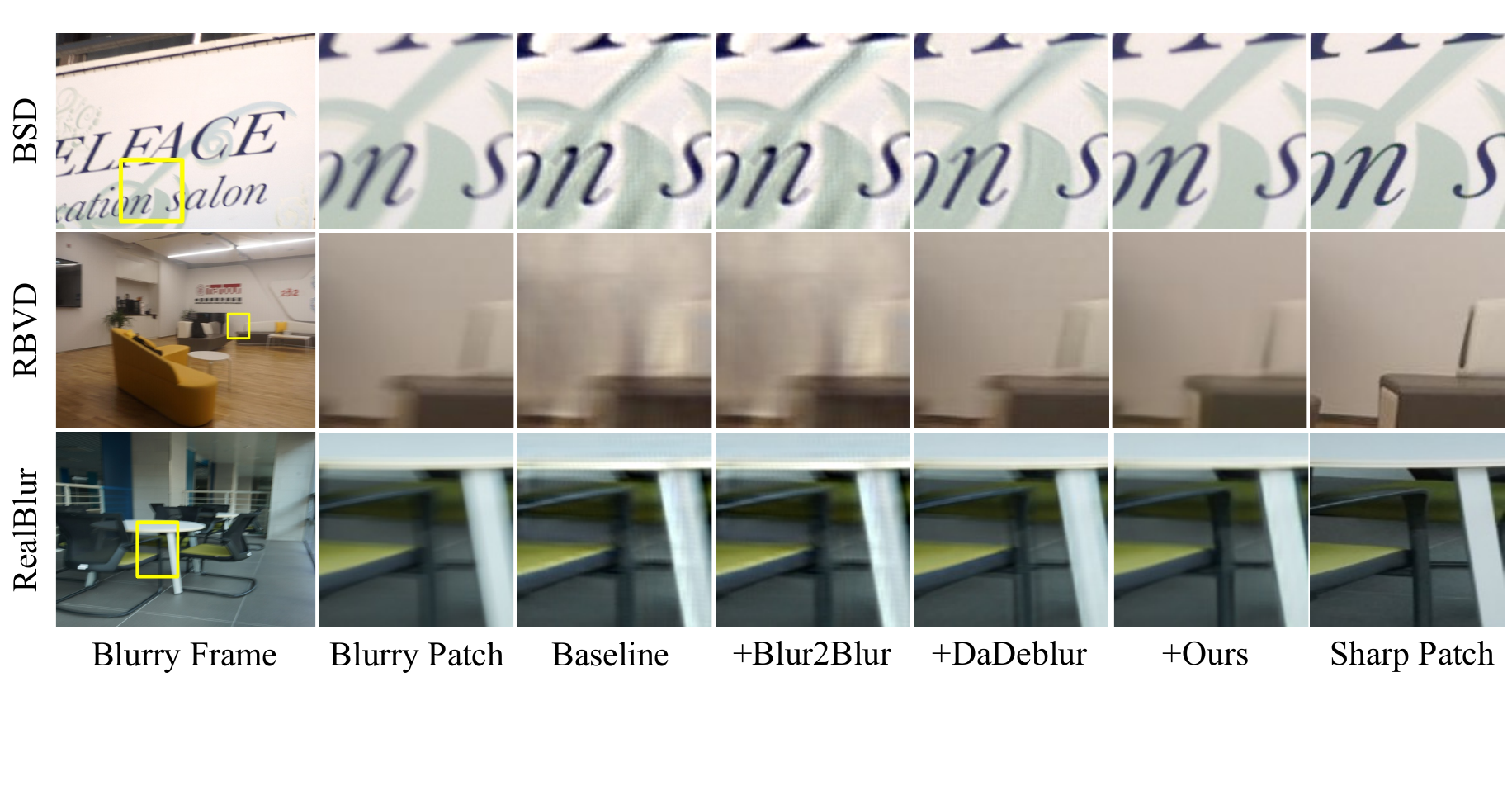}
  \vspace{-2mm}
  \caption{Qualitative results on BSD~\cite{zhong2020efficient}, RealBlur~\cite{rim2020real}, and RBVD~\cite{chao2022} datasets.}
  \label{fig:qualitative_online}
\end{figure*}

\begin{table*}[t]
\parbox{.36\linewidth}{
\centering
\caption{Effect of SEVD and SCSCM.}
\vspace{-2mm}
\setlength{\tabcolsep}{2.5mm}
\begin{tabular}{ccc}
\toprule
SEVD & SCSCM & PSNR$\uparrow$ / SSIM$\uparrow$ \\ 
\midrule
\usym{2717}     & \usym{2717}      & 35.61 / 0.9263 \\
\usym{2717}     & \usym{2713}      & 37.09 / 0.9342 \\
\usym{2713}     & \usym{2717}      & 36.67 / 0.9290 \\
\usym{2713}     & \usym{2713}      & \textbf{37.44} / \textbf{0.9359} \\ \bottomrule
\end{tabular}
\label{tab:5.4}
}
\hfill
\parbox{.35\linewidth}{
\centering
\caption{Effect of RSCR and SIS in SEVD.}
\vspace{-2mm}
\setlength{\tabcolsep}{3mm}
\begin{tabular}{ccc}
\toprule
RSCR & SIS & PSNR$\uparrow$ / SSIM$\uparrow$ \\ \midrule
\usym{2717}     & \usym{2717}      & 35.04 / 0.9165 \\
\usym{2717}     & \usym{2713}      & 35.11 / 0.9165 \\
\usym{2713}     & \usym{2717}      & 36.27 / 0.9302 \\
\usym{2713}     & \usym{2713}      & \textbf{37.44} / \textbf{0.9359} \\ \bottomrule
\end{tabular}
\label{tab:5.5}
}
\hfill
\parbox{.28\linewidth}{
\centering
\caption{Effect of \#update of $\mathcal{D}_e$ parameters in SCSCM.}
\vspace{-2mm}
\setlength{\tabcolsep}{2.5mm}
\begin{tabular}{cc}
\toprule
\#Update & PSNR$\uparrow$ / SSIM$\uparrow$ \\ \midrule
1     & 36.00 / 0.9284 \\
2     & 37.44 / \textbf{0.9359} \\
3     & \textbf{37.57} / 0.9348 \\ \bottomrule
\end{tabular}
\label{tab:5.6}
}
\vspace{-4mm}
\end{table*}

%% file: sec/5_ablation.tex
\section{Ablation studies}
\label{sec:ablation}
To verify the effectiveness of each component, we conducted ablation studies with SelfHVD$_{BasicVSR++}$. Additional ablation studies are provided in Sec.~\textcolor{red}{D}
of the supplementary material, including analyses on mask design, SEVD and SCSCM module, optical flow models, sharp frame selection intervals, and supervised pre-training.\\
\textbf{Effect of SEVD and SCSCM.} To evaluate the individual and combined contributions of Self-Enhanced Video Deblurring (SEVD) and Self-Constrained Spatial Consistency Maintenance (SCSCM), we conduct an ablation study with four configurations, as shown in Table~\ref{tab:5.4}. The results demonstrate that introducing either SEVD or SCSCM independently leads to noticeable improvements in both PSNR and SSIM. Notably, the integration of both modules yields the highest performance, indicating their complementary effects in enhancing video deblurring quality.\\
\textbf{Effect of RSCR and SIS in SEVD.} We conduct an ablation study to evaluate the effectiveness of Residual Sharp Clue Removal (RSCR) and Supervision Information Selection (SIS), as shown in Table~\ref{tab:5.5}. RSCR brings improvement compared to randomly removing frames, confirming its targeted design is effective. SIS further improves performance by selecting the better supervision between aligned sharp frames and restored frames. \\
\textbf{Effect of \#Update of $D_e$ Parameters.} To further analyze the impact of \#update of $D_e$ parameters in SCSCM, we conducted additional experiments with varying \#update, as shown in Table~\ref{tab:5.6}. The results indicate that increasing the \#update from 1 to 2 brings a substantial improvement in both PSNR and SSIM. However, further increasing \#update to 3 yields only marginal gains. We finally set the \#update of $D_e$ parameters to be 2.

%% file: sec/6_conclusion.tex
\section{Conclusion}
\label{sec:conclusion}
Based on the observation that sharp clues frequently appear in blurry videos captured by handheld smartphones, we propose a self-supervised handheld video deblurring method, called SelfHVD. First, we extract sharp clues as supervision. Subsequently, Self-Enhanced Video Deblurring (SEVD) leverages the model’s existing deblurring ability to construct higher-quality and diverse paired training data, while Self-Constrained Spatial Consistency Maintenance (SCSCM) ensures the spatial alignment between input and output. Finally, we construct a synthetic dataset GoProShake and a real-world dataset HVD collected with a smartphone. Extensive experiments on these two datasets and the common real-world datasets demonstrate that SelfHVD significantly outperforms existing self-supervised ones.

%% file: sec/7_acknowledgements.tex
\section*{Acknowledgements}
\label{sec:acknowledgements}
{\raggedright
This work was partially supported by the National Natural Science Foundation of China under Grant No.~62476067, the China Postdoctoral Science Foundation under Grant No.~2025M784371, and the OPPO Research Fund.\par
}

%% file: sec/X_suppl.tex
\clearpage
\setcounter{page}{1}
\maketitlesupplementary

\renewcommand{\thesection}{\Alph{section}}
\renewcommand{\thetable}{\Alph{table}}
\renewcommand{\thefigure}{\Alph{figure}}
\setcounter{section}{0}
\setcounter{figure}{0}
\setcounter{table}{0}

\noindent The content of the appendix involves: \vspace{0.7em}

\begin{itemize}[leftmargin=1.5em, label=--, itemsep=0.7em]
  \item Presence of sharp frames in Appendix~\ref{sec:stable}.
  \item More details of GoProShake and HVD in Appendix~\ref{sec:datasets}.
  \item Effectiveness on object motion blur in Appendix~\ref{sec:dynamic}.  
  \item More ablation studies in Appendix~\ref{sec:moreab}.
  \item More evaluation details and results in Appendix~\ref{sec:er}.
\end{itemize}

\section{Presence of Sharp Frames}
\label{sec:stable}
\begin{figure*}
  \centering
  \includegraphics[width=0.99\linewidth]{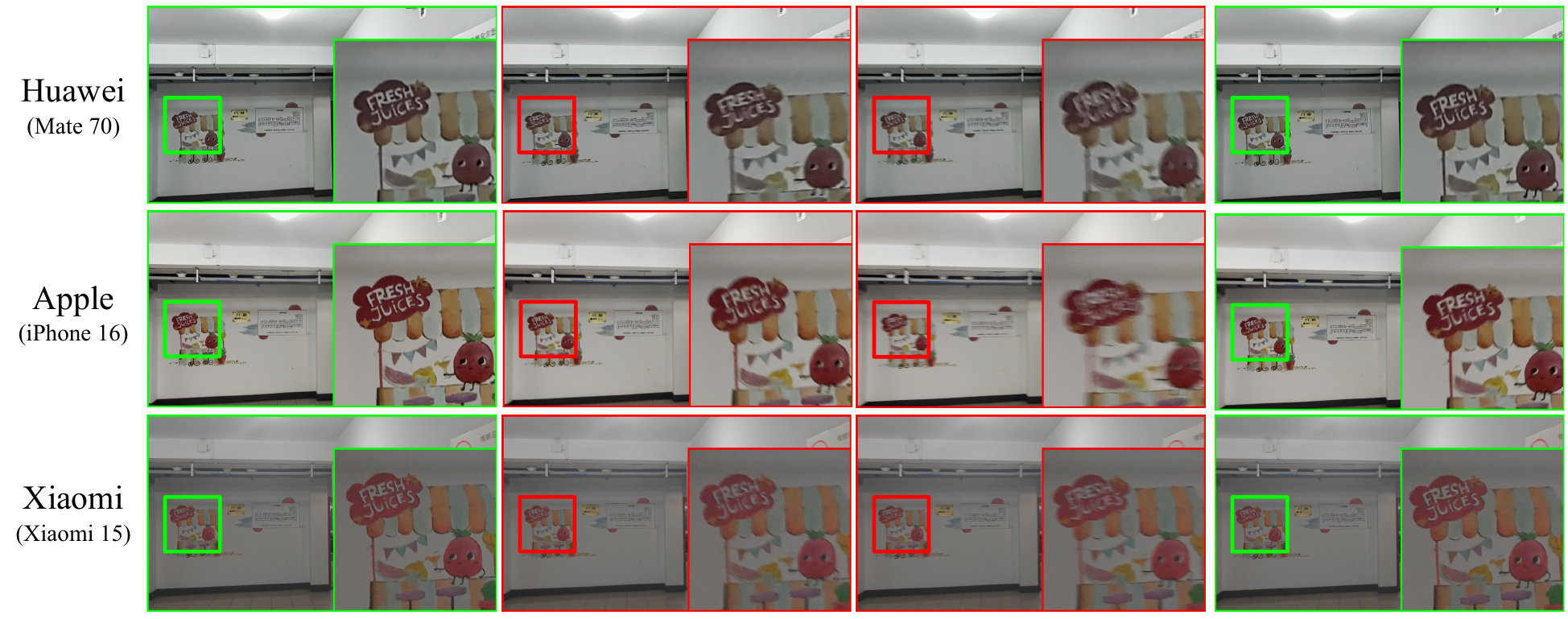}
  \vspace{-2mm}
  \caption{Illustration of sharp and blurry frames coexisting in videos captured by different smartphones (Huawei, Apple, and Xiaomi) during handheld walking. Each row corresponds to a different device. Green boxes highlight sharp frames, while red boxes indicate blurry frames.}
  \vspace{-4mm}
  \label{fig:phone}
\end{figure*}

Modern smartphones from almost all major manufacturers, such as Huawei, Xiaomi and Apple, are equipped with image stabilization technologies like OIS as standard features. In typical handheld scenarios, such as walking or jogging while recording, the shake frequency commonly falls within the effective compensation range of these stabilization systems. As a result, sharp frames consistently appear in handheld video recordings, as illustrated in Fig.~\ref{fig:phone} and further supported by the sharp frame ratios reported in Table~\ref{tab:recent_phone_ois_clarity}, which are typically around 30\% across different models from various manufacturers. This observation forms the foundation for our self-supervised approach.

\begin{table}[h]
\centering
\caption{Recent smartphone models (2022–2024) with OIS support and sharp frame ratio in captured videos}
\vspace{-2mm}
\resizebox{1.\linewidth}{!}{
\begin{tabular}{ccccc}
\toprule
Brand & Model & Release Year & OIS & Sharp Frame Ratio \\
\midrule
\multirow{3}{*}{Huawei} 
  & Mate 50           & 2022 & \usym{2713} & 33.33\% \\
  & Mate 60           & 2023 & \usym{2713} & 32.50\% \\
  & Mate 70           & 2024 & \usym{2713} & 38.85\% \\
\hline
\multirow{3}{*}{Xiaomi} 
  & Xiaomi 13         & 2022 & \usym{2713} & 34.34\% \\
  & Xiaomi 14         & 2023 & \usym{2713} & 31.91\% \\
  & Xiaomi 15         & 2024 & \usym{2713} & 33.68\% \\
\hline
\multirow{3}{*}{Apple}   
  & iPhone 14         & 2022 & \usym{2713} & 30.00\% \\
  & iPhone 15         & 2023 & \usym{2713} & 32.55\% \\
  & iPhone 16         & 2024 & \usym{2713} & 33.11\% \\
\bottomrule
\end{tabular}
}
\label{tab:recent_phone_ois_clarity}
\vspace{-4mm}
\end{table}

\section{More Details of GoProShake and HVD}
\label{sec:datasets}
\subsection{Synthetic Dataset GoProShake}
\label{sec:goproshake}
GoPro \cite{Nah_2017_CVPR} chooses to record the sharp information to be integrated over time for blur image generation, which can be formulated as: 
\begin{equation}
    \mathbf B=g\left(\frac{1}{T}\int_{t=0}^T\mathbf S(t)dt\right)\simeq g\left(\frac{1}{K}\sum_{i=0}^{K-1}\mathbf S[i]\right)
\label{equ:gopro}
\end{equation}
where $T$ represent the exposure time and $\mathbf S(t)$ denote the sensor signal of a sharp image at time $t$. Similarly, $K$ denotes the number of sampled frames and $\mathbf S[i]$ represents the signal of the $i$-th sharp frame captured during the exposure. The function $g$ is the camera response function (CRF) that maps the latent sharp signal $\mathbf S(t)$ to an observed image and is approximated with a gamma curve:
\begin{equation}
    g(x)=x^{1/\gamma}
\label{equ:gamma}
\end{equation}
where $\gamma$ is commonly set to 2.2. 

Different from GoPro \cite{Nah_2017_CVPR}, the synthesis process of GoProShake considers the OIS technology in the mobile phone. According to \textbf{Characteristic of Handheld Video} in Sec~\ref{subsec:3.1}, the blur degree is often proportional to the motion speed of the shooting device. Therefore, unlike GoPro \cite{Nah_2017_CVPR}, whose number of sampled frames $K$ in Eq.~(\ref{equ:gopro}) remains odd constant within the same video, in GoProShake, it is proportional to the motion speed between frames. Specifically, we first use MonST3R \cite{zhang2024monst3r} to roughly estimate the 3D motion trajectory of the mobile phone, and then calculate the movement distance between frames based on the motion trajectory:
\begin{equation}
    d_i=\int(v_r+v_s)dt
\label{equ:distance}
\end{equation}
where $v_r$ and $v_s$ represent the rotational and translational velocity vector of the mobile phone, respectively. From the pose obtained by MonST3R \cite{zhang2024monst3r}, we can calculate the rotational and translational distances vector between frames, then we can get the rotational velocity vector $v_r$ and translational velocity vector $v_s$ from the distances vector. 

\begin{figure*}[t]
    \centering
    \begin{overpic}[percent,width=0.99\linewidth]{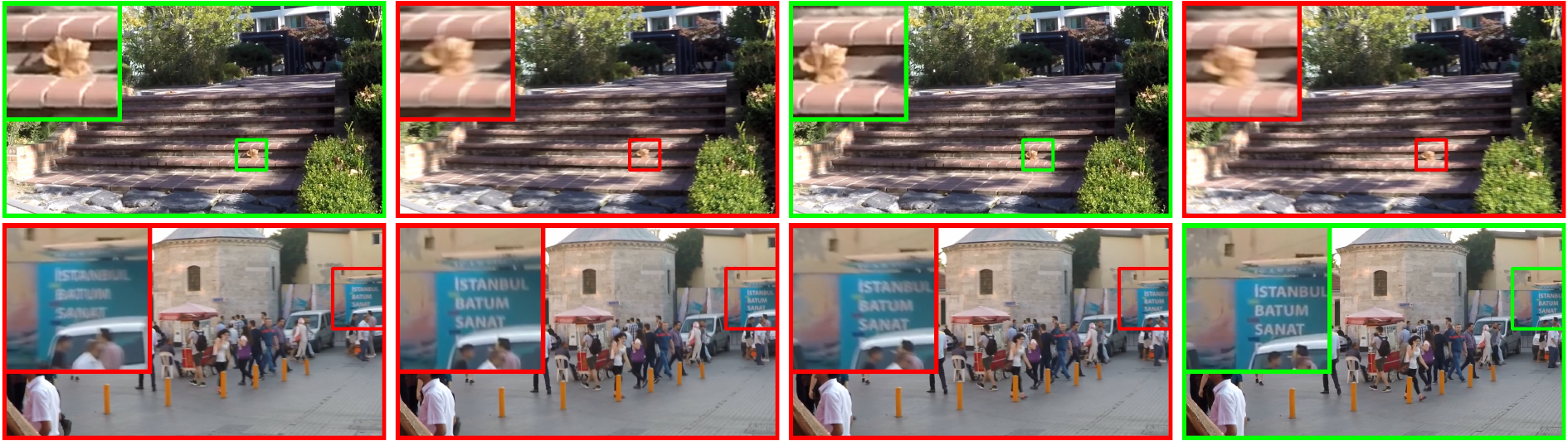}
    \end{overpic}
    \vspace{-2mm}
    \caption{Visualization of GoProShake dataset. The top and bottom are training and test videos, respectively. GoProShake takes into account the OIS technology on handheld video capture, synthesizing blurry videos (red boxes) that contain sharp frames (green boxes).}
    \label{fig:goproshake}
    \vspace{-2mm}
\end{figure*}

Our synthesis process also uses a sliding window approach, with the window size and step size set to the same value as GoPro \cite{Nah_2017_CVPR}, which is the number of sampled frames $K$. Therefore, the sequence number $m_j$ of the middle frame of the $j$-th sliding window is:
\begin{equation}
    m_j=j * K  +K / 2
\label{equ:index}
\end{equation}
where $j=0,1,2...$ and we can calculate the average movement distance in the $j$-th sliding window as:
\begin{equation}
    \bar{d}_j=\frac{1}{K}\sum_{i=m_j-K/2}^{m_j+K/2}d_i
\label{equ:meandistance}
\end{equation}
Then the number of sampled frames for each sliding window is:
\begin{equation}
    k_j = \min\left(1, K * \frac{\bar{d}_{j}}{D} \right)
\label{equ:kj}
\end{equation}
where $D$ is a normalization constant. The number of sampled frames $k_j$ is proportional to the movement distance $\bar{d}_j$. The smaller the movement distance, the fewer frames are sampled. The final synthetic frame $B_j$ in the $j$-th sliding window is:
\begin{equation}
    \mathbf B_j=g\left(\frac{1}{k_j}\sum_{i=m_j-k_j/2}^{m_j+k_j/2}\mathbf S[i]\right)
\label{equ:bj}
\end{equation}
where interpolation processing is applied before averaging and $g$ is the estimated CRF in \cite{Nah_2019_CVPR_Workshops_REDS}. It is noted that the $j$-th frame is sharp when $k_j=1$. Our GoProShake dataset contains 22 training videos and 11 test videos, consistent with GoPro \cite{Nah_2017_CVPR}. The visualization of the video from GoProShake as shown in Fig.~\ref{fig:goproshake}

\subsection{Real-World Dataset HVD}
\label{sec:hvd}
\begin{figure*}[t]
    \centering
    \begin{overpic}[percent,width=0.99\linewidth]{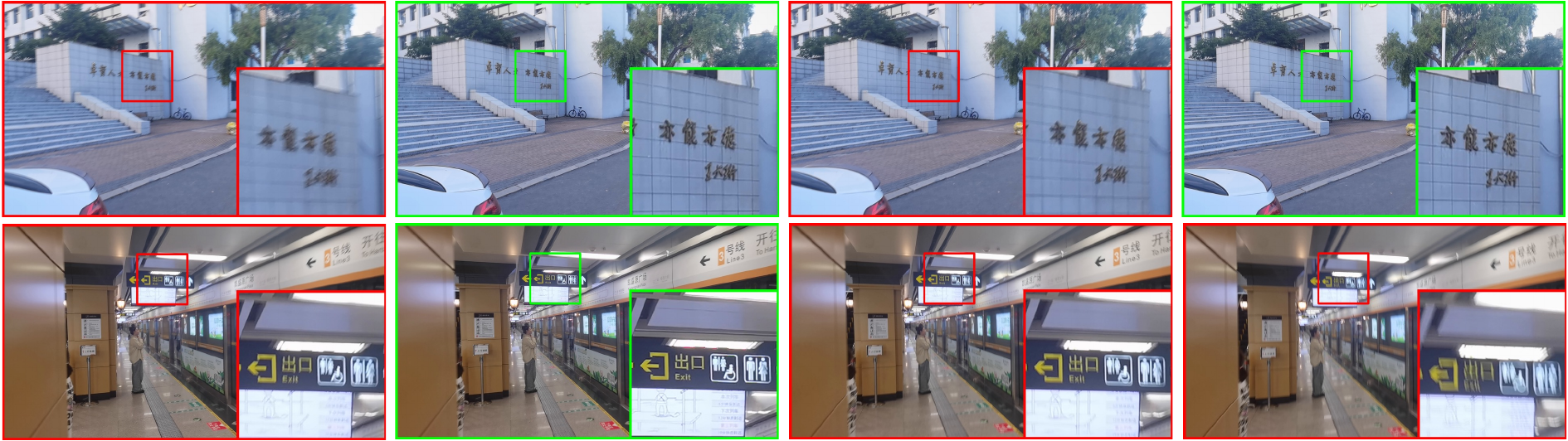}
    \end{overpic}
    \vspace{-2mm}
    \caption{Visualization of HVD dataset. Sharp frames (green boxes) are present and reliable in most cases of handheld shooting scenarios.}
    \label{fig:hvd}
\end{figure*}

The videos of HVD are captured by walking normally in various scenes, such as \textit{night scenes of commercial streets, campuses, underground parking lots} and \textit{subway stations}, using HUAWEI P40, Xiaomi 15 and iPhone 16. All videos are recorded at a frame rate of 30fps with an exposure time of 16ms. HVD contains a total of 180 videos, with 120 used for training and 60 (20 Huawei P40, 20 Xiaomi 15, and 20 iPhone 16) for testing. The visualization of the video from HVD as shown in Fig.~1(a) and Fig.~\ref{fig:hvd}.

\begin{table*}[t!]
\centering
\caption{SEVD and SCSCM ablation across backbones (RVRT~\cite{liang2022rvrt} and BasicVSR++~\cite{chan2022basicvsrpp}) and datasets (GoProShake and HVD).}
\vspace{-2mm}
\begin{tabular}{cc|cc|cc}
\toprule
SEVD & SCSCM
& \makecell[c]{\textbf{RVRT} \\ \textbf{on GoProShake} \\ PSNR / SSIM}
& \makecell[c]{\textbf{RVRT} \\ \textbf{on HVD} \\ MUSIQ / MANIQA}
& \makecell[c]{\textbf{BasicVSR++} \\ \textbf{on GoProShake} \\ PSNR / SSIM} 
& \makecell[c]{\textbf{BasicVSR++} \\ \textbf{on HVD} \\ MUSIQ / MANIQA} \\ 
\midrule
\usym{2717}     & \usym{2717} & 34.34 / 0.9155           & 26.9798 / 0.2098          & 35.61 / 0.9263           & 26.9677 / 0.2060 \\
\usym{2717}     & \usym{2713} & 36.11 / 0.9288           & 27.6052 / 0.2189          & 37.09 / 0.9342           & 27.7905 / 0.2103 \\
\usym{2713}     & \usym{2717} & 35.89 / 0.9210           & 27.2834 / 0.2149          & 36.67 / 0.9290           & 27.2445 / 0.2061 \\ 
\usym{2713}     & \usym{2713} & \textbf{36.31} / \textbf{0.9300} & \textbf{28.4142} / \textbf{0.2627} & \textbf{37.44} / \textbf{0.9359} & \textbf{28.0040} / \textbf{0.2175} \\ 
\bottomrule
\end{tabular}
\label{tab:combined_simple}
\vspace{-4mm}
\end{table*}

\section{Deblurring Effects on Object Motion Blur}
\label{sec:dynamic}
Our method is capable of handling not only camera motion blur but also object motion blur. As described in the main paper, due to object motion is typically non-uniform, relatively sharp content is retained when the object is still or moves slowly. Video deblurring models can aggregate information across multiple frames, allowing these sharp contents to provide crucial clues for dealing with blur when the object moves fast. Compared to deblurring results from the blurry video without sharp clues, those from the original blurry video perform better on object motion, as illustrated in the middle row of Fig.~\ref{fig:sevd} in the main paper and Fig.~\ref{fig:dynamic_sup}. The high-quality training pairs constructed by SEVD also offer more reliable supervision for object motion deblurring. As a result, our method achieves better than DaDeblur~\cite{He2024DADeblur} on the object motion blur. Some visualizations on HVD can be seen in Fig.~\ref{fig:dynamic}.

\section{More Ablation Studies}
\label{sec:moreab}
\subsection{Effect of the Masks}
\label{sec:mask}
Table~\ref{tab:mask} shows the ablation results of the $\mathbf{M}_{uncer}$ and the $\mathbf{M}_{occ}$. Both masks individually bring performance gains, indicating their effectiveness in handling uncertain or occluded regions. And combining both yields the best results, highlighting their complementary roles in enhancing reconstruction quality. Fig.~\ref{fig:mask} and Fig.~\ref{fig:mask2} visualize the proposed masks on the synthetic dataset GoProShake and the real-world dataset HVD, respectively. As shown in the figures, the masks effectively identify and suppresses misaligned regions that result from inaccurate optical flow or large content discrepancies between frames. This prevents erroneous supervision and ensures that only reliable regions contribute to the learning process.

\begin{table}[]
\centering
\caption{Effect of $\mathbf M_{uncer}$ and $\mathbf M_{occ}$.}
\vspace{-2mm}
\begin{tabular}{cc|cc}
\hline
$M_{uncer}$ & $M_{occ}$ & PSNR$\uparrow$ & SSIM$\uparrow$ \\ \hline
\usym{2717}     & \usym{2717}      & 36.13              & 0.9157 \\
\usym{2717}     & \usym{2713}      & 37.25              & 0.9334 \\
\usym{2713}     & \usym{2717}      & 37.00              & 0.9343 \\
\usym{2713}     & \usym{2713}      & \textbf{37.44}     & \textbf{0.9359} \\ \hline
\end{tabular}
\label{tab:mask}
\vspace{-4mm}
\end{table}

\subsection{Effect of SEVD and SCSCM}
\label{sec:backbones&datasets} 
Table~\ref{tab:combined_simple} presents the ablation results of SEVD and SCSCM on two backbones (RVRT~\cite{liang2022rvrt} and BasicVSR++~\cite{chan2022basicvsrpp}) across both the synthetic dataset GoProShake and the real-world dataset HVD. Individually introducing SEVD or SCSCM improves performance across most metrics, validating their respective contributions. Notably, the combination of SEVD and SCSCM consistently achieves the best performance in all settings, highlighting their complementary effectiveness across different backbones and datasets.

\subsection{Effect of the Optical Flow Model} 
\label{sec:op}
\begin{figure*}[t]
    \centering
    \includegraphics[width=1\linewidth]{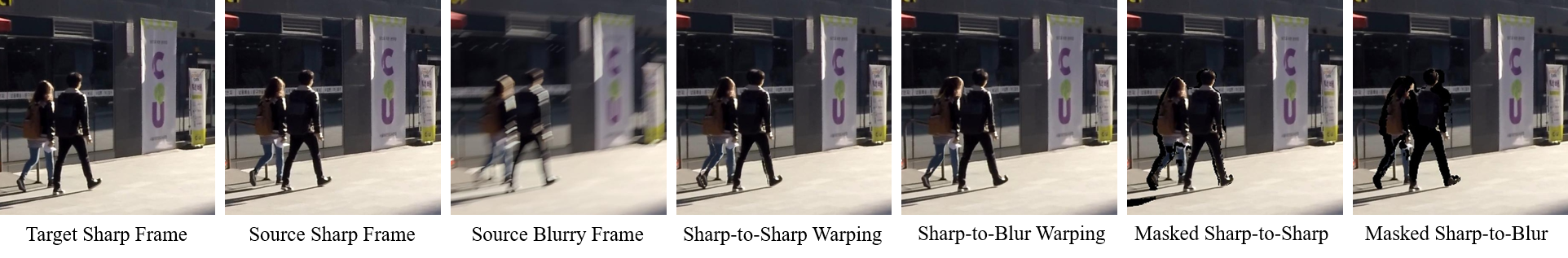}
    \vspace{-6mm}
    \caption{%
        Qualitative alignment results under different blur levels. The source sharp and blurry frames are generated by fusing different numbers of high-frame-rate images, with the sharp frame typically being a single mid-frame and the blurry frame formed by averaging multiple consecutive frames.
    }
    \label{fig:level}
    \vspace{-4mm}
\end{figure*}

\begin{figure}[t]
    \centering
    \begin{overpic}[percent,width=0.99\linewidth]{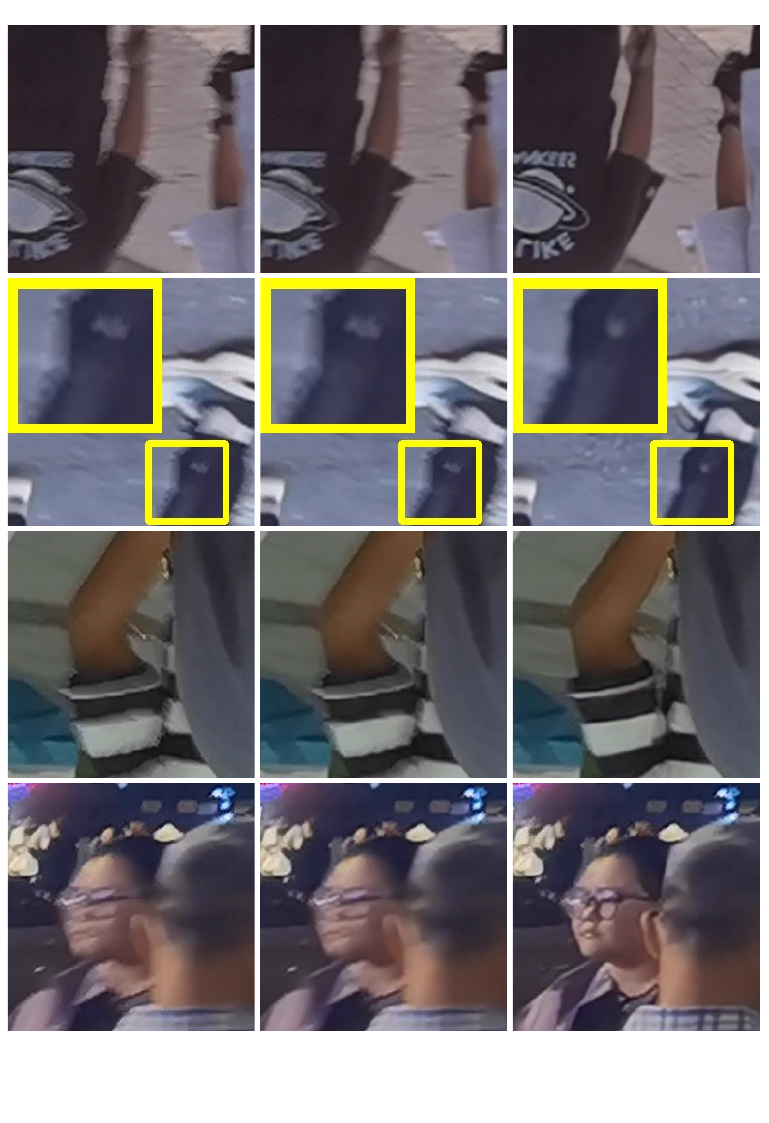}
    \put(11,4.5){$\tilde{\mathbf{B}}_i$}
    \put(28.5,4.5){$\mathcal{D}(\tilde{\mathbf{B}}; \mathbf{\Theta}_{\mathcal{D}})_i$}
    \put(50.5,4.5){$\mathcal{D}(\mathbf{B}; \mathbf{\Theta}_{\mathcal{D}})_k$}
    \end{overpic}
    \vspace{-6mm}
    \caption{
    From left to right: sharp-clues-less blurry video, deblurring result of sharp-clues-less blurry video, deblurring result of original input video. SEVD improves object motion blur handling by constructing higher-quality paired data.}
    \label{fig:dynamic_sup}
\end{figure}

\begin{figure}[t]
    \centering
    \begin{overpic}[percent,width=0.99\linewidth]{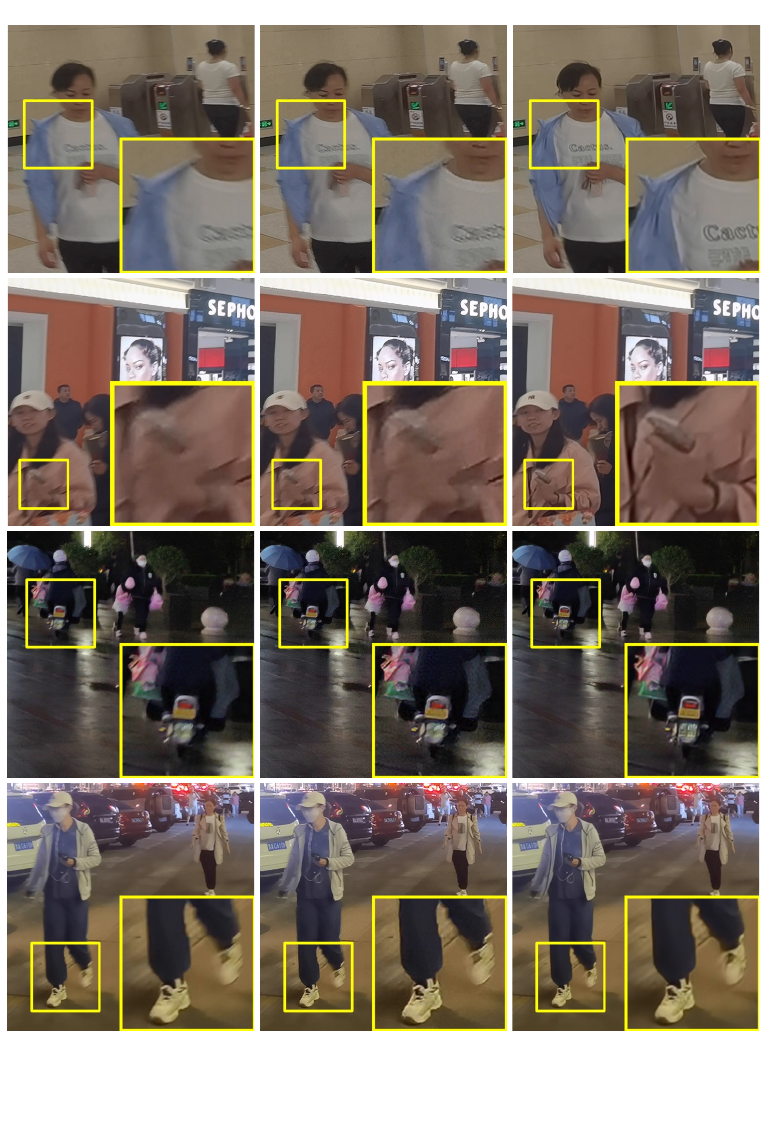}
    \put(11,4.5){Blur}
    \put(29,4.5){DaDeblur}
    \put(55,4.5){Ours}
    \end{overpic}
    \vspace{-6mm}
    \caption{Visual comparisons on handling object motion blur. Even for challenging cases involving object motion blur, our method achieves better performance compared to DaDeblur.}
    \label{fig:dynamic}
    \vspace{-2mm}
\end{figure}

We investigate the impact of different optical flow models on our deblurring performance. As shown in Table~\ref{tab:flow}, replacing SEA-RAFT~\cite{wang2024sea} with RAFT~\cite{teed2020raft} or FlowFormer++\cite{shi2023flowformer++} results in PSNR drops of 0.66dB and 0.20dB, respectively, and slight SSIM declines. This demonstrates that SEA-RAFT provides more accurate optical flow estimation, enabling better frame alignment and enhanced restoration quality. Furthermore, Fig.\ref{fig:level} illustrates the robustness of SEA-RAFT under varying degrees of blur, where it consistently yields reliable flow predictions even in severely degraded regions.

\begin{table}[t]
\centering
\caption{Effect of the optical flow method.}
\vspace{-2mm}
\begin{tabular}{lcc}
\toprule
Optical Flow Method & PSNR$\uparrow$ & SSIM$\uparrow$ \\
\midrule
RAFT     & 36.78 & 0.9353 \\
FlowFormer++   & 37.24 & 0.9327 \\
SEA-RAFT   & \textbf{37.44} & \textbf{0.9359} \\
\bottomrule
\end{tabular}
\label{tab:flow}
\end{table}

\subsection{Effect of Sharp Frame Selection Interval} 
\label{sec:sharp_interval}
Table~\ref{tab:sharp_interval} investigates how different sharp frame selection intervals affect selection accuracy and deblurring performance on GoProShake. The accuracy is computed by comparing our selected sharp frames with manually labeled ones. A smaller interval 5 yields denser supervision but lower accuracy 71.03\%. Increasing the interval enhances selection accuracy, reaching 98.28\% at interval 30, but this comes at the cost of sparsity in supervision. Among all settings, an interval of 20 achieves the best deblurring performance, striking a good balance between selection reliability and coverage. Therefore, we adopt this interval in all main experiments. We also apply the same sharp frame selection strategy to the real-world HVD dataset, and observe reasonable accuracy 91.88\%, confirming that real videos also contain sharp frames that can serve as reliable supervision.

\begin{table}[t]
    \centering
    \caption{Selection accuracy and deblurring performance under different sharp frame selection intervals on GoProShake. An interval of $k$ means one sharp frame is selected for every $k$ frames.}
    \vspace{-2mm}
    \begin{tabular}{ccc}
        \toprule
        \makecell{Selection\\Interval} & \makecell{Selection\\Accuracy} & PSNR$\uparrow$/SSIM$\uparrow$ \\
        \midrule
        5   & 71.03\% & 37.33 / 0.9305 \\
        10  & 88.51\% & 37.36 / 0.9342 \\
        20  & 96.77\% & \textbf{37.44} / \textbf{0.9359} \\
        30  & \textbf{98.28}\% & 37.03 / 0.9326 \\
        \bottomrule
    \end{tabular}
    \label{tab:sharp_interval}
    \vspace{-2mm}
\end{table}

\subsection{Effect of Supervised Pre-training.}
\label{subsec:ss}
To assess the effect of supervised pre-training, we first evaluate the performance of the fully supervised BasicVSR++ trained on different datasets, as shown in the upper part of Table~\ref{tab:pretrain}. These results serve as baselines. We then apply our self-supervised method, SelfHVD\(_{\textit{BasicVSR++}}\), to fine-tune these pre-trained models on our real-world dataset HVD. The lower part of the table presents the results after self-supervised adaptation. Regardless of the pre-training dataset, SelfHVD\(_{\textit{BasicVSR++}}\) consistently improves quality over the original supervised models. And better supervised pre-training generally results in better performance after self-supervised fine-tuning. These results demonstrate that our self-supervised method effectively enhances the performance of different pre-trained models on real-world handheld blurry videos.

\begin{table}[]
\centering
\caption{Quantitative comparison on real-world HVD dataset. `Pre-training' denotes the dataset used for pre-training models. The best results in each category are \textbf{bolded}, and the second-best results are \underline{underlined}.}
\vspace{-2mm}
\setlength{\tabcolsep}{2.3pt}
\scalebox{0.85}{
\begin{tabular}{ccccc}
\toprule
\multicolumn{2}{c}{Methods}                                                      & Pre-training & MUSIQ$\uparrow$ & MANIQA$\uparrow$   \\ 
\midrule
\multirow{3}{*}{\begin{tabular}[c]{@{}c@{}c@{}}Fully-\\Super-\\vised\end{tabular}} & \multirow{3}{*}{BasicVSR++} & BSD-2-16 & \textbf{26.5821} & \textbf{0.2303}  \\
                                                                             &  & GoPro & \underline{25.9927} & \underline{0.2014}  \\
                                                                             &  & GoProShake & 25.2488 & 0.2006  \\ \midrule
\multirow{4}{*}{\begin{tabular}[c]{@{}c@{}c@{}}Self-\\Super-\\vised\end{tabular}}  & \multirow{4}{*}{SelfHVD$_{BasicVSR++}$} & - & 28.0040 & 0.2175 \\
                                                                             & & BSD-2-16 & \underline{28.1463} & 0.2135  \\
                                                                             & & GoPro & 27.7622 & \underline{0.2189}  \\
                                                                             & & GoProShake & \textbf{28.2212} & \textbf{0.2231}  \\
                                                                            \bottomrule
\end{tabular}
}
\label{tab:pretrain}
\end{table}

\begin{table}[t]
\centering
\caption{Model complexity and average inference time comparison of different video deblurring backbones.}
\vspace{-2mm}
\setlength{\tabcolsep}{1mm}
\begin{tabular}{lccc}
\hline
\multicolumn{1}{c}{Networks}                                & \#Params(M) & \#FLOPs(G) & Time(ms) \\ \hline
IFIRNN~\cite{nah2019recurrent}       & 1.64        & 29.55      & 16.53    \\
ESTRNN~\cite{zhong2020efficient}     & 2.47        & 146.96     & 79.31    \\
RVRT~\cite{liang2022rvrt}            & 10.78       & 1379.84    & 472.50   \\
BasicVSR++~\cite{chan2022basicvsrpp} & 9.76        & 72.53      & 27.38    \\ \hline
\end{tabular}
\label{tab:re}
\end{table}

\begin{table}[t]
\centering
\caption{Temporal consistency comparison of self-supervised methods on the synthetic dataset GoProShake. The best results in each category are \textbf{bolded}, and the second-best results are \underline{underlined}.}
\vspace{-2mm}
\resizebox{1.\linewidth}{!}{
\begin{tabular}{lcccc}
\hline
\multicolumn{1}{c}{Methods} &
\multicolumn{1}{c}{tOF$\downarrow$} &
\multicolumn{1}{c}{tLP$\downarrow$} &
\multicolumn{1}{c}{FVD$\downarrow$} &
\multicolumn{1}{c}{VBench$\uparrow$} \\
\hline
Ren~\etal~\cite{ren_deblur}                 & 4.9773 & 3.8688 & 112.60 & 0.8978 \\
DaDeblur~\cite{He2024DADeblur}             & 2.0680 & 2.2800 & 31.40  & 0.9018 \\
SelfHVD$_{IFIRNN}$                          & \textbf{1.5423} & 1.5953 & \textbf{6.18} & 0.9064 \\
SelfHVD$_{ESTRNN}$                          & 1.7895 & 1.8452 & 7.24   & 0.9044 \\
SelfHVD$_{RVRT}$                            & 1.7451 & \underline{1.4712} & 6.85   & \underline{0.9065} \\
SelfHVD$_{BasicVSR++}$                      & \underline{1.5911} & \textbf{1.1539} & \underline{6.71}   & \textbf{0.9069} \\
\hline
\end{tabular}
}
\label{tab:tc}
\end{table}

\section{More Evaluation Details and Results}
\label{sec:er}
\subsection{More Evaluation Details}
Under full-supervision, the GoProShake training set (w/ GT) is used for training, while the GoProShake and HVD-Huawei test sets, as well as the HVD-Xiaomi and HVD-iPhone, are used for evaluation. Under self-supervision, for synthetic data, the GoProShake training set (w/o GT) is used for training and the GoProShake test set is used for evaluation; for real-world data, the HVD-Huawei training set is used for training and the HVD-Huawei test set, HVD-Xiaomi, and HVD-iPhone are used for evaluation.

\subsection{More Visual Results}
\label{sec:visual}
\begin{figure*}
    \centering
    \includegraphics[width=0.99\linewidth]{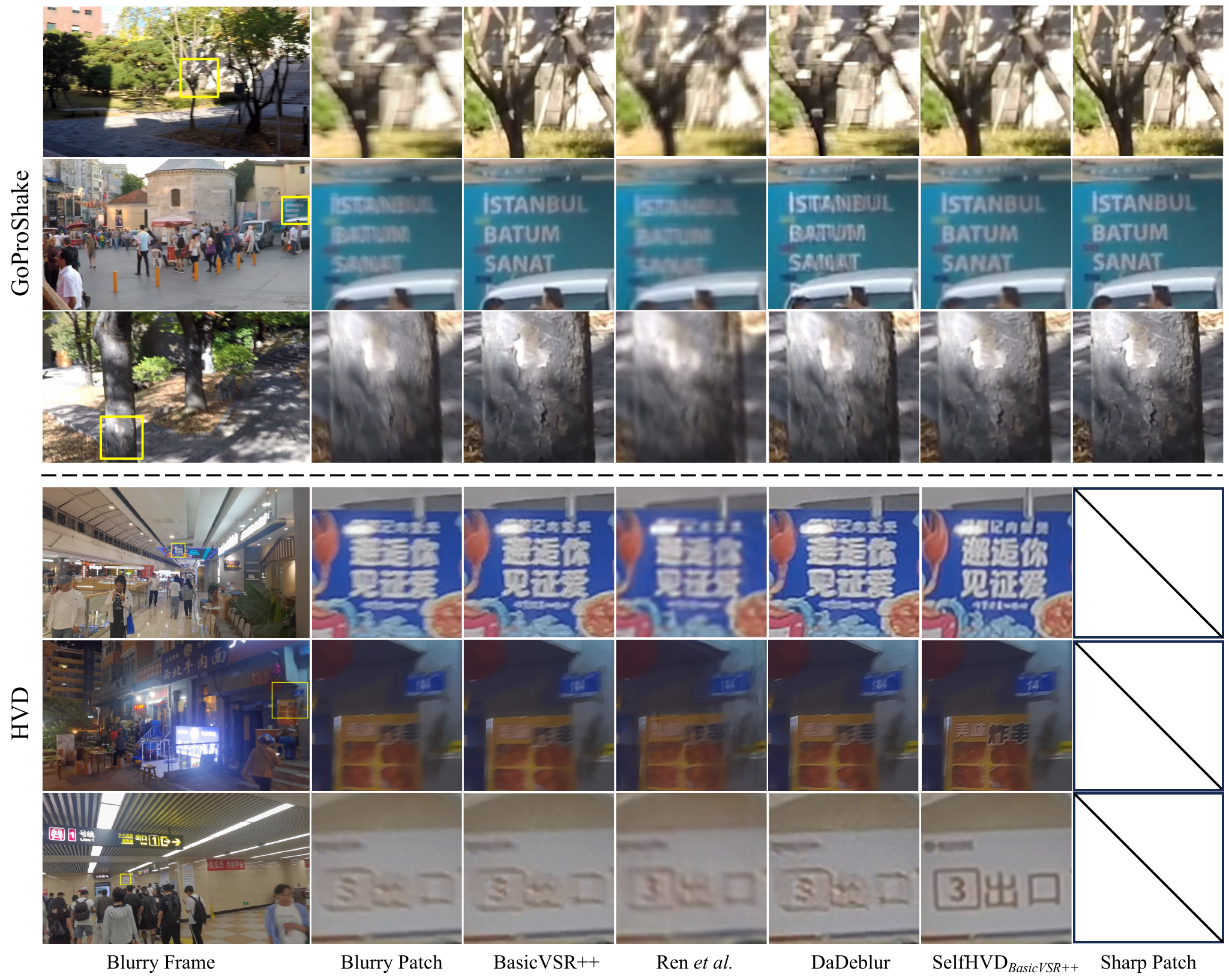}
    \caption{More qualitative comparison on our synthetic dataset GoProshake and real-world dataset HVD.}
    \label{fig:qualitative_hvd}
\end{figure*}

\begin{figure*}
    \centering
    \includegraphics[width=0.99\linewidth]{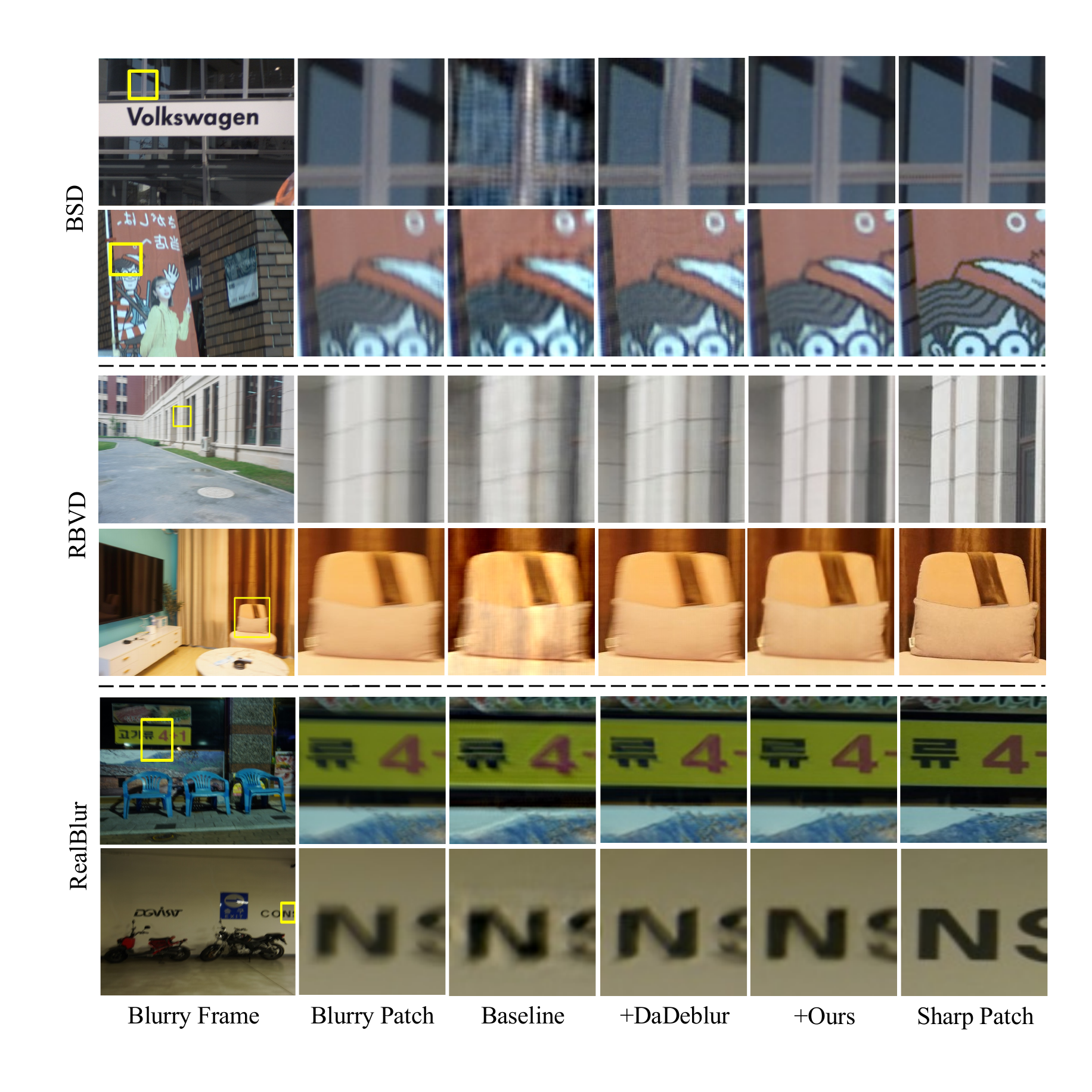}
    
    \caption{More qualitative comparison on BSD, RVRB and RealBlur.}
    \label{fig:online2}
\end{figure*}

\begin{figure*}[t]
    \centering
    \includegraphics[width=0.9\linewidth]{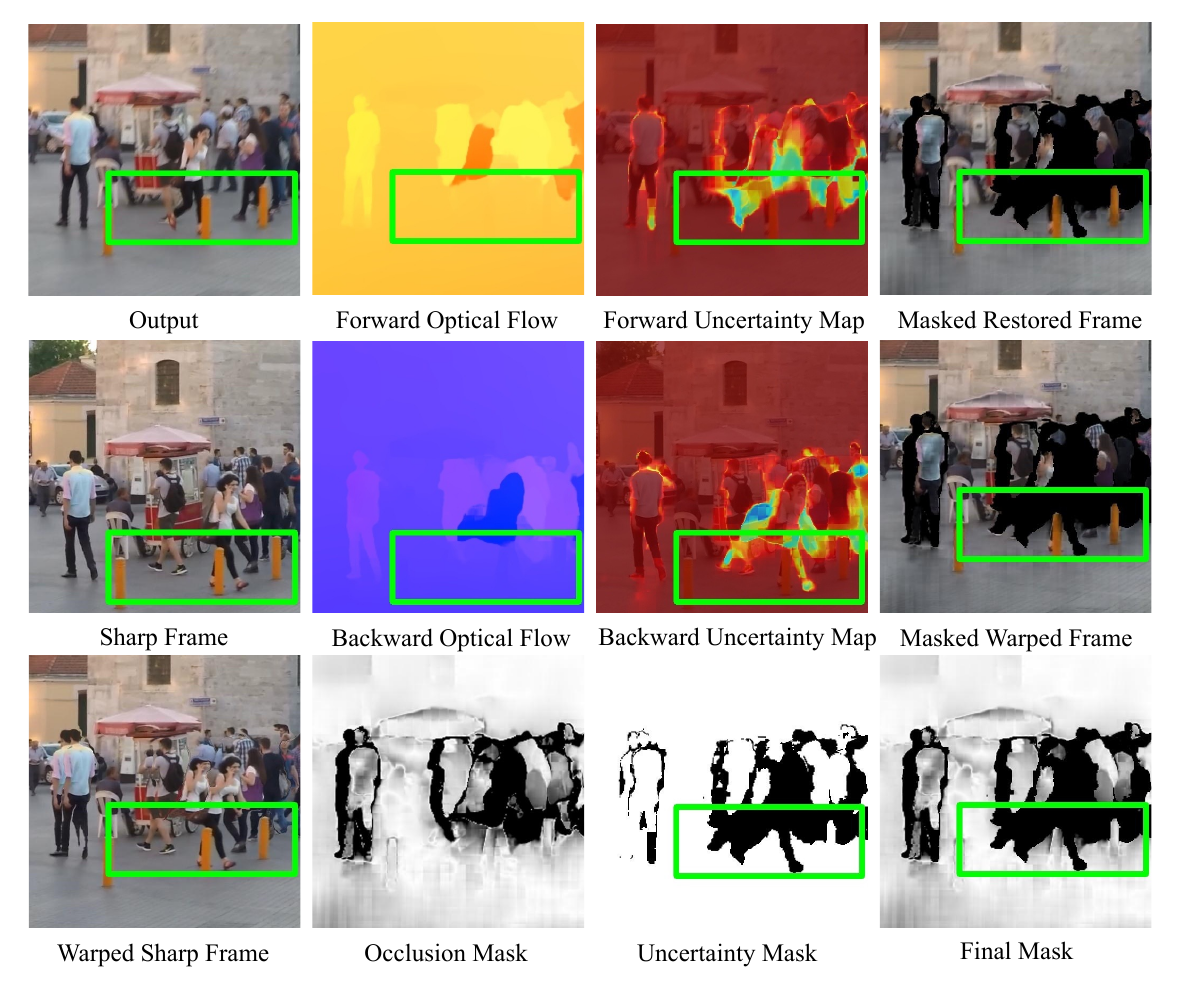}
    \vspace{-6mm}
    \caption{Visualization of the masks on the synthetic dataset GoProShake. The green box indicates the region where the optical flow is inaccurate. The uncertainty map will perceive the inaccurate region, and the uncertainty mask is calculated from it to mask the region.
    }
    \label{fig:mask}
\end{figure*}

\begin{figure*}[t]
    \centering
    \includegraphics[width=0.73\linewidth]{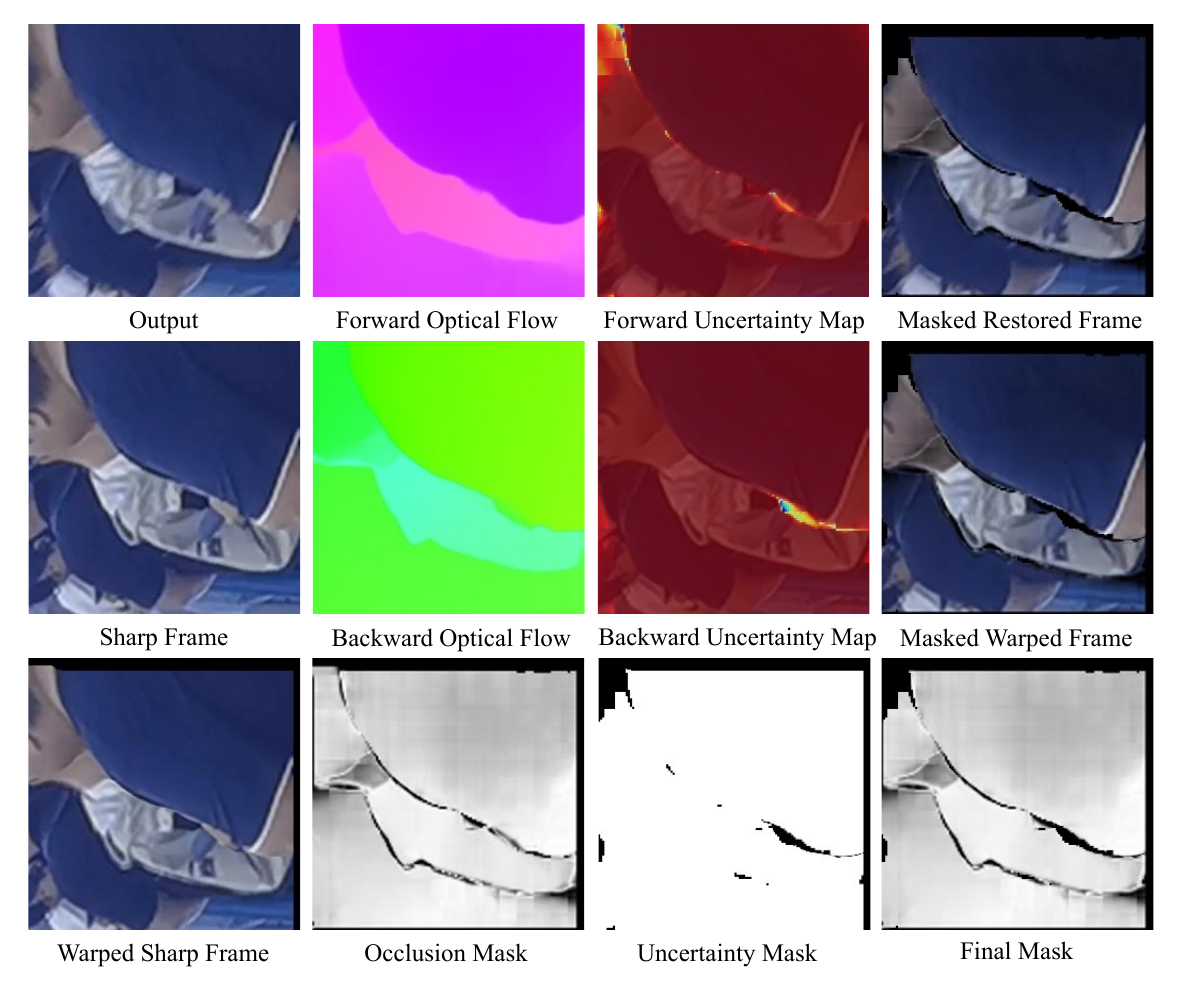}
    
    \vspace{-4mm}
    
    \includegraphics[width=0.73\linewidth]{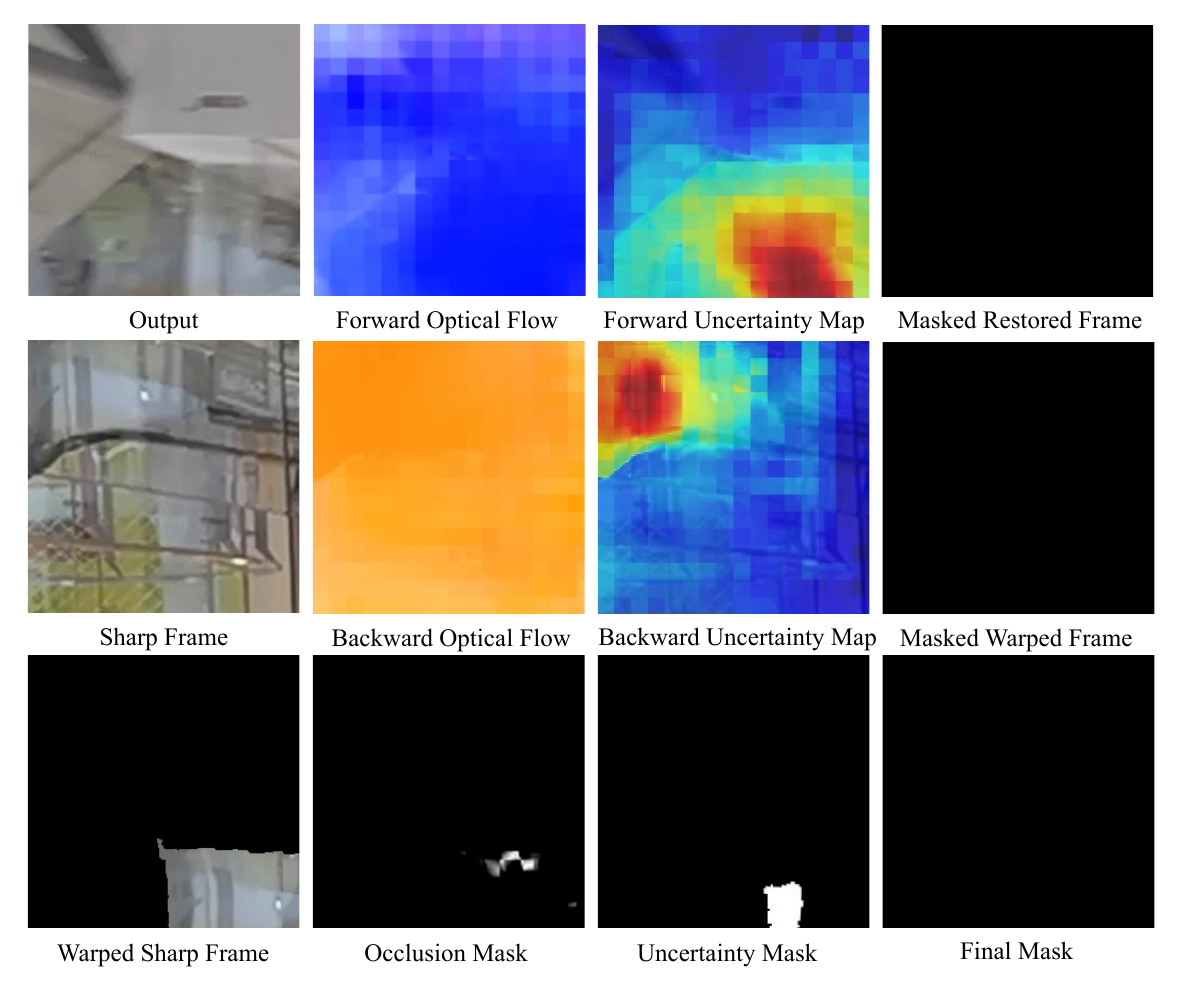}
    
    \vspace{-4mm}
    
    \caption{More mask visualization on the real-world dataset HVD, showing the behavior of our masks under varying degrees of content discrepancy between the predicted output and the sharp frame.}
    \label{fig:mask2}
\end{figure*}

To further validate the visual effectiveness of our method, we present additional qualitative comparisons in Figs.\ref{fig:qualitative_hvd} and \ref{fig:online2}. As shown in Fig.\ref{fig:qualitative_hvd}, SelfHVD$_{\text{BasicVSR++}}$ consistently generates sharper results on our synthetic dataset GoProShake, outperforming previous self-supervised approaches, and illustrates the robustness of our method on the real-world dataset HVD. Lastly, Fig.\ref{fig:online2} demonstrates that under the same test-time training setting as DaDeblur, SelfHVD achieves better visual quality on BSD~\cite{zhong2020efficient}, RBVD~\cite{chao2022}, and RealBlur~\cite{rim2020real}. These results further support the quantitative improvements reported in the main paper and confirm the generalization capability of SelfHVD across both synthetic and real-world datasets.

\subsection{Running Efficiency}
\label{sec:re}
Since our framework can be applied to various video deblurring networks, we select representative backbones with different architectures and model sizes, including IFIRNN~\cite{nah2019recurrent}, ESTRNN~\cite{zhong2020efficient}, RVRT~\cite{liang2022rvrt}, and BasicVSR++~\cite{chan2022basicvsrpp}, where ESTRNN~\cite{zhong2020efficient} is also adopted by Ren~\etal~\cite{ren_deblur} and DaDeblur~\cite{He2024DADeblur}. The model complexity (numbers of parameters (\#Param), FLOPs (\#FLOPs)), and the average inference time (Time) are shown in Tab.~\ref{tab:re}.

\subsection{Temporal Consistency}
\label{sec:tc}
We adopt tOF and tLP~\cite{chu2020learning} as temporal consistency metrics. As shown in Tab.~\ref{tab:tc}, SelfHVD built on different backbones (IFIRNN~\cite{nah2019recurrent}, ESTRNN~\cite{zhong2020efficient}, RVRT~\cite{liang2022rvrt}, and BasicVSR++~\cite{chan2022basicvsrpp}) consistently achieves lower tOF and tLP values on GoProShake than previous self-supervised methods Ren~\etal~\cite{ren_deblur} and DaDeblur~\cite{He2024DADeblur}. These results demonstrate the better temporal consistency of SelfHVD.

